\documentclass[]{article}
\usepackage{authblk}

\usepackage[utf8]{inputenc} 
\usepackage[T1]{fontenc}    
\usepackage{hyperref}       
\usepackage{url}            
\usepackage{booktabs}       
\usepackage{amsfonts}       
\usepackage{nicefrac}       
\usepackage{microtype}      
\usepackage{fullpage}

\usepackage{hyperref}
\usepackage{adjustbox}
\usepackage{url}
\usepackage{graphicx}
\usepackage{amsmath}
\usepackage{bbm}
\usepackage{paralist}
\usepackage{Definitions}
\usepackage[noend]{algorithmic}
\usepackage[vlined,ruled]{algorithm2e}
\usepackage{color}
\usepackage{enumitem}
\usepackage{subfig}
\usepackage{graphicx}
\usepackage{tabularx}
\usepackage{mathtools}
\usepackage[dvipsnames]{xcolor}

\newcommand{\redqueen}{{\textsc{Red\-Queen}}\xspace}
\newcommand{\cheshire}{{\textsc{Che\-shire}}\xspace}

\newcommand{\xhdr}[1]{\vspace{1mm}\noindent{{\bf #1. }}}

\newcommand\blfootnote[1]{%
  \begingroup
  \renewcommand\thefootnote{}\footnote{#1}%
  \addtocounter{footnote}{-1}%
  \endgroup
}

\graphicspath{{./FIG/}}

\title{\cheshire: An Online Algorithm for Activity \mbox{Maximization in Social Networks}}

\author[1]{Ali Zarezade$^{*}$}
\author[2]{Abir De$^{*}$}
\author[1]{Hamid R. Rabiee}
\author[3]{Manuel Gomez-Rodriguez}

\affil[1]{Sharif University, zarezade@ce.sharif.edu}
\affil[2]{IIT Kharagpur, abir.de@cse.iitkgp.ernet.in}
\affil[3]{Max Planck Institute for Software Systems, manuelgr@mpi-sws.org}

\date{}

\begin{document}


\maketitle

\begin{abstract}
User engagement in social networks depends critically on the number of online \emph{actions} their users take in the 
network. Can we design an algorithm that finds when to incentivize users to take actions to maximize the overall 
activity in a social network?
In this paper, we model the number of online actions over time using multidimensional Hawkes processes, derive an 
alternate representation of these processes based on stochastic differential equations (SDEs) with jumps and, exploiting 
this alternate representation, address the above question from the perspective of stochastic optimal control of SDEs with 
jumps.
We find that the optimal level of incentivized actions depends linearly on the current level of overall actions. Moreover, the 
coefficients of this linear relationship can be found by solving a matrix Riccati differential equation, which can be solved 
efficiently, and a first order differential equation, which has a closed form solution.
As a result, we are able to design an efficient online algorithm, \cheshire, to sample the optimal times of the users'{} incentivized
actions.
Experiments on both synthetic and real data gathered from Twitter show that our algorithm is able to consistently maximize
the number of online actions more effectively than the state of the art.
\end{abstract}

\blfootnote{$^{*}$\scriptsize Authors contributed equally. This work was done during Ali Zarezade'{}s
and Abir De'{}s internships at Max Planck Institute for Software Systems.}

\section{Introduction}
\label{sec:intro}
People are constantly performing a wide variety of online actions in a growing number of online platforms such as social networking 
sites, question answering (Q\&A) sites or wikis.
In most of these platforms, these actions can be \emph{exogenous}, taken by users at their own initiative, or \emph{endogenous} 
actions, taken by users as a \emph{response} to previous actions by other users.
For example, 
in social networking sites, users post small pieces of information, which can then trigger likes, shares or replies by other 
users;
in Q\&A sites, users can ask questions, which are then answered and curated by other users; 
or, in wikis, editors write content, which is later reviewed and refined by other editors in a collaborative fashion.
In this context, a natural question emerges: how much should we incentivize a small number of users to take more 
initiatives to drive the overall number of actions to a certain level (\eg, at least twice actions per day per user)?

The ``activity shaping" problem was first studied by Farajtabar et al.~\cite{Farajtabar2015}, who derived a time dependent linear 
relation between the \emph{intensity} of exogenous actions and the overall intensity of actions in a social network under a model 
of actions based on multivariate Hawkes processes and, exploiting this connection, developed a convex optimization framework 
for activity shaping.
One of the main shortcomings of their framework is that it provides deterministic exogenous in\-ten\-si\-ties that do not adapt to 
changes in the users'{} intensities and, as a consequence, it is less effective than our proposed algorithm, as shown in 
Section~\ref{sec:experiments}.
More recently, Farajtabar et al.~\cite{farajtabar2016msc} developed a heuristic method that splits the time window of interest 
into stages and adapts to changes in the users'{} intensities at the beginning of each stage. However, their method is suboptimal, 
it does not have \emph{provable} guarantees, and it achieves lower performance than our method.

In this paper, we design a novel online algorithm for the ``activity maximization" problem in social networks, one of the most 
important instances of the activity shaping problem, in which the goal is to maximize the number of actions in the 
network.\footnote{Our methodology can be easily extended to online platforms without an explicit underlying network between 
users. However, for ease of exposition, we focus on social networking sites.}
More in detail, we represent users'{} actions using the framework of temporal point processes, which characterizes the continuous time 
interval between actions using conditional intensity functions~\cite{Aalen2008}, and model endogenous and exogenous actions using 
multidimensional Hawkes processes~\cite{Hawkes1971}. 
Then, we derive an alternate representation of multidimensional Hawkes processes using SDEs with 
jumps and, exploiting this alternate representation, cast the activity maximization problem as a novel optimal control problem for SDEs 
with jumps. 
Our problem formulation differs from the traditional control literature~\cite{hanson2007} in two key technical aspects, which have not been 
considered until very recently~\cite{redqueen17wsdm}:
\begin{itemize}[leftmargin=1cm]
\item[I.] The control signal, used to sample the times of incentivized actions, is a multidimensional conditional intensity 
while previous work considered the control signal to be a time-varying real vector.
\item[II.] The users'{} intensities are stochastic Markov processes and thus the dynamics are doubly stochastic. In contrast, 
previous work considered deterministic (typically constant) intensities.
\end{itemize}
Moreover, we find that the optimal level of incentivized actions depends linearly on the current level of overall actions. Moreover, 
the coefficients of this linear relationship can be found by solving a matrix Riccati differential equation, which can be solved using 
many well-known efficient numerical solvers~\cite{garrett2013}, and a first order differential equation, which has closed form 
solution.
This allows for an efficient and relatively simple online procedure to sample the optimal times of the users'{} incentivized 
actions, which can be implemented in a few lines of code (refer to Algorithm~\ref{alg:sampling}).
Finally, we perform experiments on both synthetic and real data gathered from Twitter and show that our method is able to consistently maximize 
the number of actions more effectively than the state of the art~\cite{Farajtabar2015,farajtabar2016msc}.

\subsection{Further related work}
In addition to the paucity of work on the activity shaping problem~\cite{Farajtabar2015,farajtabar2016msc}, discussed previously, 
our work also relates to previous work on stochastic optimal control, the influence maximization problem, and temporal point processes.

In the traditional control literature~\cite{hanson2007}, two key aspects of our problem formulation---intensities as control signals and stochastic intensities---have 
been large understudied.
Only very recently, Zarezade et al.~\cite{redqueen17wsdm} and Wang et al.~\cite{wang2017} have considered these aspects, however, our approach differs in 
several ways:
The work by Zarezade et al. is more closely related to ours, however, they solve a different problem, the when-to-post problem~\cite{spasojevic2015post}, where
one aims to optimize a social network user'{}{}s broadcasting strategy to capture the greatest attention from the followers. 
Such problem is significantly less challenging than activity maximization---it only considers $1$-hop information propagation---and, as a consequence, their resulting 
sampling algorithm is very different to ours. 
The work by Wang et al. focuses on two different problems, the when-to-post problem and opinion control, their strategy is open-loop 
and their control policy depends on the expectation of the uncontrolled dynamics, which needs to be calculated approximately by a 
time consuming sampling process.
In contrast, our framework is closed-loop, our control policy only depends on the current state of the dynamics and the feedback coefficients only need to be calculated 
once off-line.

The influence maximization problem~\cite{CheWanWan2010,Du2013,Kempe2003,RicDom02} aims to find a set of nodes in a social network whose 
initial adoption of a certain idea or product can trigger the largest expected number of follow-ups. However, previous work on the influence maximization 
problem differs from our work in two main aspects.
First, in influence maximization, the state of each user is often assumed to be binary, either adopting a product or not.
However, such assumption does not capture the recurrent nature of product usage, where the frequency of the usage matters.
Second, while influence maximization methods identify a set of users to provide incentives, they do not typically provide a quantitative prescription on how 
much incentive should be provided to each user.

Temporal point processes has been increasingly used for representation and modeling in a wide range of applications in social and information 
systems, \eg, information propagation~\cite{Rodriguez2011, Du2013, zhao2015seismic}, opinion dynamics~\cite{de2016learning}, product 
competition~\cite{Valera2015}, information reliability~\cite{reliability2017tabibian}, or human learning~\cite{hdhp2017learning}.
However, in such context, there is still a paucity of algorithms based on stochastic optimal control of temporal point processes~\cite{redqueen17wsdm, wang2017}.

\section{Preliminaries}
\label{sec:background}
In this section, we first revise the framework of temporal point processes~\cite{Aalen2008} and then describe how to use such
framework to model endogenous and exogenous actions in social networks~\cite{Farajtabar2014}.

\subsection{Temporal point processes}
A univariate temporal point process is a stochastic process whose realization consists of a sequence of discrete events localized in time, $\Hcal = \{t_i\in \RR_{+} \,\vert\, i\in \mathbb{N}_+,\, t_i<t_{i+1} \}$.
%
%
A univariate temporal point process can be equivalently represented by a counting process $N(t)$, which counts the number of events before time $t$, \ie, 
\begin{equation*}
N(t) = \sum_{t_i \in \Hcal} u(t - t_i), 
\end{equation*}
where $u(t) = 1$ if $t \geq 0$ and $u(t) = 0$ otherwise.
%
%
Then, we can characterize the counting process using the conditional intensity function $\lambda^{*}(t)$, which is the 
conditional probability of observing an event in an infinitesimal window $[t, t + dt)$ given the history of event times up to time $t$, $\Hcal(t) = \{t_i \in \Hcal \,\vert\, t _i < t \}$, \ie,
\begin{equation*}
\lambda^{*}(t) \, dt = \PP\cbr{\text{event in $[t, t+dt) \,\vert\, \Hcal(t)$}} = \EE[dN(t) \,\vert\, \Hcal(t)],
\end{equation*}
where $dN(t) := N(t+dt) - N(t) \in \{0, 1\}$, the sign $^{*}$ means that the intensity may depend on the history $\Hcal(t)$, and the functional form for the intensity is often 
designed to capture the phenomena of interest. 
Moreover, given a function $f(t)$, it will be useful to define the convolution with respect to $dN(t)$ as
%
%
\begin{equation*} 
f(t) \star dN(t) := \int_{0}^{t} f(t - s) dN(s) = \sum_{t_i \in \Hcal(t)} f(t - t_i). 
\end{equation*}
One can readily extend the above definitions to multivariate (or multidimensional) temporal point processes, which have been recently used to represent many different types of event data 
produced in social networks, such as the times of tweets~\cite{smart16}, retweets~\cite{zhao2015seismic} or links~\cite{ farajtabar2015coevolve}.
More specifically, a realization of an $m$-dimensional temporal point process consists of $m$ sequences of discrete events localized in time, $\Hcal = \cup_{u \in [m]} \Hcal_u$, where 
%
%
$\Hcal_u = \{t_i | t_i \in \RR_{+} \,\vert\, i\in \mathbb{N}_+,\, t_i<t_{i+1} \}$, and it can be represented by an $m$-dimensional counting process $\mathbf{N}(t)$, where $N_u(t)$ counts the 
number of events in the $u$-th sequence before time $t$. 
Similarly, such counting process can be characterized by $m$ intensity functions, \ie,
\begin{equation*}
\lambdab^{*}(t)dt = \EE[d\Nb(t)|\Hcal(t)], 
\end{equation*}
where $\lambdab^{*}(t) = (\lambda_1^{*}(t), 
\ldots, \lambda_m^{*}(t))$ and
%
%
 $\Hcal(t) = \{t_i \in \Hcal | t _i < t \}$, and, given a function $f(t)$, one can define the convolution with respect to $d\Nb(t)$ as
%
%
\begin{equation*}
\int_{0}^{t} f(t - s) d\Nb(s) = ( \sum_{t_i \in \Hcal_u(t)} f(t - t_i) )_{u \in [m]}.
\end{equation*}

Next, we use the above background to revisit how to jointly model endogenous and exogenous actions in social networks~\cite{Farajtabar2014} and then derive an alternate model representation based on SDEs
with jumps, which will be useful to design our stochastic optimal control algorithm.

\subsection{Modeling endogenous and exogenous actions in social networks}
Given a directed network $\mathcal{G}=(\mathcal{V},\mathcal{E})$ with $|\mathcal{V}|=n$ users, we model both endogenous and exogenous actions taken by all the users using an 
$m$-dimensional Hawkes process $\bm{N}(t)$, where $N_u(t)$ counts the number of actions taken by user $u$ before time $t$.
More specifically, the intensities of this process are given by~\cite{Farajtabar2014, Hawkes1971}:
\begin{align} \label{eq:multi-hawkes}
	\lambdab^{*}(t) = \bm{\mu}_0 +  \bm{A} \int_0^t \kappa(t-s) \, d\bm{N}(s), 
\end{align}
where the first term models exogenous actions---actions users take at their own initiative---and the second term model endogenous actions---actions users take as response to the actions taken by their neighbors within the network. 
Here, we parametrize the strength of \emph{influence} between users using a sparse nonnegative \emph{influence matrix} $\bm{A} = (a_{u v{}}) \in \mathbb{R}_{+}^{m\times m}$, where $a_{u v{}}$
means user u'{}s actions directly triggers \emph{follow-ups} from user $v{}$, and $\kappa(t) = e^{-\omega t} \, u(t)$ is an exponential kernel modeling the decay of influence over time.
Note that the second term makes the intensities dependent on the history and thus a stochastic process by itself.
In the remainder of the paper, we drop the sign $^{*}$ from the intensities for the notational convenience.

The following alternative representation of the above process will be useful to design our stochastic optimal control algorithm for the activity maximization (proven in Appendix~\ref{app:hawkes-dynamics}):
\begin{proposition} \label{prop:hawkes}
	Let $\bm{N}(t)$ be an $m$-dimensional counting process with an associated intensity $\lambdab(t)$ given by Eq.~\ref{eq:multi-hawkes}. Then, the tuple $(\bm{N}(t), \lambdab(t))$ is a doubly stochastic 
	Markov process, whose dynamics can be defined by the following jump SDEs:
%
%
	\begin{align} \label{eq:hawkes-dyn}
		d\bm{\lambda}(t) = \left[w \bm{\mu}_0 - w \bm{\lambda}(t) \right] dt + \bm{A} \, d\bm{N}(t), 
	\end{align}
	with the initial condition $\lambdab(0)= \lambdab_0$.
\end{proposition}


\section{Problem Formulation}
\label{sec:formulation}
In this section, we first describe a mechanism to steer endogenous actions in social networks and then formally state the online activity maximization 
problem for such mechanism.

\subsection{Steering endogenous actions}
We steer endogenous actions (or events) in the counting process $\bm{N}(t)$ by introducing a counting process $\bm{M}(t)$, with control intensities $\bm{u}(t)$, which triggers additional 
follow-ups. More specifically, this counting process modulates the corresponding intensities $\bm{\lambda}(t)$ as follows: 
\begin{align} \label{eq:steering-multi-hawkes}
	\lambdab^{*}(t) = \underset{\mbox{\scriptsize Organic actions}}{\underbrace{\bm{\mu}_0 +  \bm{A} \int_0^t \kappa(t-s) \, d\bm{N}(s)}} 
	+  \underset{\mbox{\scriptsize Incentivized actions}}{\underbrace{\bm{A} \int_0^t \kappa(t-s) \, d\bm{M}(s)}},
\end{align}
where we assume the strength of influence $\bm{A}$ between users is the same both for the organic and incentivized actions, as previous work~\cite{Farajtabar2014, farajtabar2016msc}.
Then, it is easy to derive the following alternative representation, similarly as in Proposition~\ref{prop:hawkes}, which we will use in our stochastic optimal control 
algorithm: 
\begin{proposition}\label{thm:hawkes-dynamics}
	Let $\bm{N}(t)$ be a multidimensional counting process with associated intensities $\bm{\lambda}(t)$, given by Eq.~\ref{eq:steering-multi-hawkes}, and $\bm{M}(t)$ be a controllable 
	counting process with an associated intensity $\bm{u}(t)$. Then, the system dynamics can be defined by the following jump SDEs:
	\begin{align}\label{eq:sys-dyn}
	d\bm{\lambda}(t) = \left[w \bm{\mu}_0 - w \bm{\lambda}(t) \right] dt + \bm{A} \, d\bm{N}(t) + \bm{A} \, d\bm{M}(t)
	\end{align}
	with the initial condition $\lambdab^{*}(0)= \lambdab_0$.
\end{proposition}
%

\subsection{The (online) activity maximization problem} 
Given a directed network $\mathcal{G}=(\mathcal{V},\mathcal{E})$ with $|\mathcal{V}|=n$ users, our goal is to find the optimal
control intensities $\bm{u}(t)$ that minimize the expected value of a particular loss function $\ell(\bm{\lambda}(t),\bm{u}(t))$ of the 
users'{}s organic and control intensities over a time window $(t_0,t_f]$, \ie,
\begin{align}\label{eq:prob-def}
&\underset{\bm{u}(t_0,t_f]}{\text{minimize}} \quad
\mathbb{E}_{(\bm{N},\bm{M})(t_0,t_f]}\left[ \phi(\bm{\lambda}(t_f)) + \int_{t_0}^{t_f} \ell(\bm{\lambda}(t), \bm{u}(t)) \, dt \right] \nonumber \\
&\text{subject to} \quad u_i(t) \geq 0, \,\, \forall t \in (t_0,t_f], \, i=1,\ldots,n
\end{align}
where $\bm{u}(t_0,t_f]$ denotes the control intensity functions from $t_0$ to $t_f$, the dynamics of $\bm{N}(t)$ are given by Eq.~\ref{eq:sys-dyn}, 
and the expectation is taken over all possible realizations of the two counting process $\bm{N}(t)$ and $\bm{M}(t)$ during interval $(t_0,t_f]$.
Here, by considering a loss that is nonincreasing with respect to the organic intensities $\bm{\lambda}(t)$, we will penalize low organic intensities, 
and, by considering a loss that is nondecreasing with respect to the control intensities $\bm{u}(t)$, we will limit the number of posts we steer.
Finally, note that the optimal intensities $\bm{u}(t)$ at time $t$ may depend on the organic intensities $\bm{\lambda}(t)$ and thus the associated 
counting process $\bm{M}(t)$ may be doubly stochastic.

\section{Stochastic Optimal Control Algorithm}
\label{sec:method}
In this section, we tackle the activity maximization problem defined by Eq.~\ref{eq:prob-def} from the perspective of stochastic optimal control of jump SDEs~\cite{hanson2007}. 
More specifically, we first define a novel optimal cost-to-go function that accounts for the above unique aspects of our problem, show that the Bellman'{}s principle of 
optimality~\cite{bertsekas1995dynamic} still follows, and finally derive and solve the Hamilton-Jacobi-Bellman (HJB) equation to find the optimal control intensity.

\begin{definition}\label{thm:cost-def}
The optimal cost-to-go is defined as the minimum of the expected value of the cost of going from the state with intensity $\bm{\lambda}(t)$ at time $t$ to the final state at 
time $t_f$. 
\begin{align}\label{eq:cost-to-go}
J(\bm{\lambda}(t),t) = 
\min_{\bm{u}(t,t_f]} \mathbb{E}_{(\bm{N},\bm{M})(t,t_f]}\left[ \phi(\bm{\lambda}(t_f)) + \int_t^{t_f} \ell(\bm{\lambda}(s), \bm{u}(s)) \, ds \right],
\end{align}
where the expectation is taken over all trajectories of the counting processes $\bm{N}$ and $\bm{M}$ in the $(t, t_f]$ interval, given the initial 
values of $\bm{\lambda}(t)$ and $\bm{u}(t)$.
\end{definition}
To find the optimal control $u(t, t_f]$ and cost-to-go $J$, we break the problem into smaller subproblems, using Bellman'{}s principle of optimality 
(proven in the Appendix \ref{app:bellman-opt-cond}):
\begin{theorem}[Bellman'{}s Principle of Optimality] \label{thm:bellman-opt-cond}
The optimal cost-to-go function defined by Eq.~\ref{eq:cost-to-go} sa\-tis\-fies the following recursive equation:
\begin{align}\label{eq:bellman-opt}
	J(\bm{\lambda}(t),t) = \min_{\bm{u}(t, t+dt]} \big\{ \mathbb{E}_{(\bm{N},\bm{M})(t,t+dt]} \left[ J(\bm{\lambda}(t+dt),t+dt)\right ] + \ell(\bm{\lambda}(t),  \bm{u}(t)) \, dt  \big\}
\end{align}	  
where the expectation is taken over all realizations of the counting processes $\bm{N}(t)$ and $\bm{M}(t)$ in the infinitesimal interval $(t,t+dt]$.   
\end{theorem}
Next, we use the Bellman'{}s principle of optimality to derive a partial differential equation on $J$, often called HJB equation. To do so, we first assume $J$ is 
continuous and then rewrite Eq.~\ref{eq:bellman-opt} as
\begin{align}
J(\bm{\lambda}(t),t) &= \min_{\bm{u}(t, t+dt]} \big\{ \mathbb{E}_{(\bm{N},\bm{M})(t,t+dt]} \left[ J(\bm{\lambda}(t),t) + dJ(\bm{\lambda}(t),t) \right] + \ell(\bm{\lambda}(t),  \bm{u}(t)) \, dt  \big\} \nonumber \\
0 &= \min_{\bm{u}(t, t+dt]} \big\{ \mathbb{E}_{(\bm{N},\bm{M})(t,t+dt]} \left[ dJ(\bm{\lambda}(t),t) \right] + \ell(\bm{\lambda}(t),  \bm{u}(t)) \, dt  \big\} \label{eq:bellman-opt-simplified}
\end{align}

Then, we differentiate $J$ with respect to time $t$ and $\lambda(t)$ according to Ito's calculus \cite{hanson2007} by the following Lemma (proven in the Appendix \ref{app:activity-diff-cost}):
\begin{theorem}\label{thm:activity-diff-cost}
The differential of the cost-to-go function $J(\bm{\lambda}(t),t)$ given by Eq.~\ref{eq:cost-to-go} is given by:
\begin{equation*}
	dJ(\bm{\lambda}(t),t) =  J_t(\bm{\lambda}(t),t) \, dt + [w \bm{\mu}_0 - w \bm{\lambda}(t)]^T \nabla_{{\bm{\lambda}}}J(\bm{\lambda}(t),t) \, dt
	+ \sum_{i=1}^n \left[J(\bm{\lambda}(t)+\bm{a}_i, t)-J(\bm{\lambda}(t), t) \right] \left[ dN_i(t) + dM_i(t)\right]
\end{equation*}
where $\bm{a}_i$ is the $i$'{}th column of $\bm{A}$, $J_t$ is derivative of $J$ with respect to $t$ and $\nabla_{\bm{\lambda}}$ is the gradient of $J$ with respect 
to $\bm{\lambda}(t)$. 
\end{theorem}
%
Next, if we plug the above equation in Eq.~\ref{eq:bellman-opt-simplified}, then use $\mathbb{E}[N_i(t)]=\lambda_i(t)\,dt$, and $\mathbb{E}[M_i(t)]=u_i(t)\,dt$, the HJB equation follows:
\begin{align}\label{eq:hjb}
	0 &= J_t(\bm{\lambda}(t),t)
	+ [w \bm{\mu}_0 - w \bm{\lambda}(t)]^T \nabla_{{\bm{\lambda}}}  J(\bm{\lambda}(t),t) + \bm{\lambda}^T(t) \, \Delta_A J
	+ \min_{\bm{u}(t)}  \ell(\bm{\lambda}(t),  \bm{u}(t)) 
	+  \bm{u}^T(t) \, \Delta_A J
\end{align}
where $\Delta_{A}J$ denotes a vector whose $i$'{}th element is given by $(\Delta_{A}J)_i = J(\bm{\lambda}(t)+\bm{a}_i, t)-J(\bm{\lambda}(t), t)$.

To solve the above equation, we need to define the loss and penalty functions, $\ell$ and $\phi$. Following the literature on stochastic optimal 
control~\cite{bertsekas1995dynamic,hanson2007}, we consider the following quadratic forms, which will turn out to be a tractable choice:
\begin{align}\label{eq:loss-func}
	\ell(\bm{\lambda}(t), \bm{u}(t)) &= -\frac{1}{2} \bm{\lambda}^T(t) \,\bm{Q}\, \bm{\lambda}(t) + \frac{1}{2} \bm{u}^T(t) \,\bm{S}\, \bm{u}(t) \nonumber \\
	\phi(\bm{\lambda}(t_f)) &= -\frac{1}{2} \bm{\lambda}^T(t_f) \,\bm{F}\, \bm{\lambda}(t_f) 
\end{align}
where $\bm{Q}$, $\bm{F}$ and $\bm{S}$ are given symmetric matrices\footnote{\scriptsize In practice, we will consider diagonal matrices.} with $q_{ij} \geq 0$, $f_{ij} \geq 0$ 
and $s_{ij} \geq 0$ for all $i, j \in [n]$.
These matrices allow us to trade-off the number of non incentivized actions, both over time and at time $t_f$, and the number of incentivized 
actions.
Under these definitions, we can find the relationship between the optimal intensity and the optimal cost by solving the
 minimization in the HJB Eq. \ref{eq:hjb}.
\begin{align} 
&\underset{\bm{u}(t)}{\text{minimize}} \quad
\bm{u}^T(t) \, \Delta_A J + \frac{1}{2} \bm{u}^T(t) \,\bm{S}\, \bm{u}(t) \nonumber \\
&\text{subject to} \quad u_i(t) \geq 0, \,\, i=1,\ldots,n \nonumber
\end{align}
By taking the differentiation with respect to $\bm{u}(t)$, the solution of the unconstrained minimization is given by:
\begin{align} \label{eq:opt-ctrl}
	\bm{u}^*(t) = - \bm{S}^{-1} \Delta_{A} J,
\end{align}
which is the same to the solution of the constraint problem, since $(\Delta_A J)_i \leq 0$, as proved in the Appendix \ref{app:positive-delta}, 
and $s_{ij} \geq 0$ by definition.
%

Then, we substitute Eq.~\ref{eq:opt-ctrl} into Eq.~\ref{eq:hjb} and find that the optimal cost $J$ needs to satisfy the following partial 
differential equation:
\begin{align}\label{eq:hjb2}
	0 = J_t(\bm{\lambda}(t),t) 
	+ [w \bm{\mu}_0 - w \bm{\lambda}(t)]^T \nabla_{{\bm{\lambda}}}J(\bm{\lambda}(t),t)
	+ \bm{\lambda}^T(t) \, \Delta_A J
	-\frac{1}{2} \bm{\lambda}^T(t) \,\bm{Q}\, \bm{\lambda}(t) 
	- \frac{1}{2} \Delta_A J^T \bm{S}^{-1} \Delta_A J  
\end{align}
with $J(\bm{\lambda}(t_f), t_f) = \phi(\bm{\lambda}(t_f))$ as the terminal condition. The following lemma provides us
with a solution to the above equation (proven in Appendix~\ref{sec:quad-proposal}):
%
\begin{lemma} \label{lem:quad-proposal}
Any solution to the nonlinear differential equation given by Eq.~\ref{eq:hjb2} can be approximated as desired by the following
quadratic form:
\begin{equation}
J(\bm{\lambda}(t), t) = f(t) + \bm{g}(t)^T \bm{\lambda}(t) + \frac{1}{2} \bm{\lambda}(t)^T \bm{H}(t) \bm{\lambda}(t). \nonumber
\end{equation}
where $\bm{g}(t)$ and $\bm{H}(t)$ can be found by solving the following differential equations:
\begin{align}
	\dot{\bm{H}}(t) &= (\omega \Ib-\Ab)^T \bm{H}(t) + \bm{H}(t) (\omega \Ib -\Ab) + \bm{H}(t) \Ab \bm{S}^{-1} \Ab^T \bm{H}(t) + \bm{Q} \nonumber \\
	\dot{\bm{g}}(t) &= [\omega \Ib-\Ab^T+\bm{H}(t) \Ab \bm{S}^{-1} \Ab^T] \bm{g}(t) - \omega  \bm{H}(t) \bm{\mu}_0 	+ \frac{1}{2} \left[ \bm{H}(t) \Ab \bm{S}^{-1} - \Ib \right] \diag(\Ab^T \bm{H}(t) \Ab) \nonumber
\end{align}
\end{lemma}
\begin{algorithm}[t]
\small
 \begin{algorithmic}[1] 
\STATE\textbf{Initialization: } \\
\STATE Compute $\bm{H}(t)$ and $\mathbf{g}(t)$ \;
\STATE $\mathbf{u}(t) \leftarrow -\bm{S}^{-1} \big[ \Ab^T (\bm{g}(t) + \bm{H}(t) \mub_0) + \frac{1}{2} \diag(\Ab^T \bm{H}(t) \Ab) \big]$ \;
\vspace{1mm}

\STATE\textbf{General subroutine: }\\
\STATE $(i, \tau) \gets Sample(\mathbf{u}(t))$;\
\STATE $(j, s) \gets NextAction(\,)$ \;
 \WHILE{$s < \tau$}
 \STATE $\lambdab_N(t) \leftarrow \Ab \mathbf{e}_j \kappa(t - s)$ \;
 \STATE $\mathbf{u}_N(t) \leftarrow -\bm{S}^{-1} \Ab^T \bm{H}(t) \lambdab_N(t)$ \;
 \STATE $(k, r) \leftarrow Sample(\mathbf{u}_N(t))$ \;
 \IF{$r < \tau$}
 	\STATE $\tau \leftarrow r$ \;
	\STATE $i \leftarrow k$ \;
 \ENDIF
 \STATE $\mathbf{u}(t) \leftarrow \mathbf{u}(t) + \mathbf{u}_N(t)$ \;
 \STATE $(j, s) \leftarrow NextAction(\,)$ \;
 \ENDWHILE
 \STATE $\lambdab_M(t) \leftarrow \Ab \mathbf{e}_i \kappa(t - \tau)$ \;
 \STATE $\mathbf{u}_M(t) \leftarrow -\bm{S}^{-1} \Ab^T \bm{H}(t) \lambdab_M(t)$ \;
 \STATE $\mathbf{u}(t) \leftarrow \mathbf{u}(t) + \mathbf{u}_M(t)$ \;
 \STATE {\bf return} $(i, \tau)$ \\
\caption{\cheshire{}: it returns user $i$ and time $\tau$ for the next incentivized action.} \label{alg:sampling}
\end{algorithmic} 
\end{algorithm}

In the above lemma, note that the first differential equation is a matrix Riccati differential equation, which can be solved using many well-known efficient numerical 
solvers~\cite{garrett2013}, and the second one is a first order differential equation which has closed form solution. Both equations are solved backward in time with 
final conditions $\bm{g}(t_f)=\bm{0}$ and $\bm{H}(t_f)=-\bm{F}$.

Finally, given the above cost-to-go function, the optimal intensity is given by the following theorem, \ie,
\begin{theorem}
The optimal intensity for the activity maximization problem defined by Eq.~\ref{eq:prob-def} with quadratic loss 
and penalty function is given by:
\begin{align}
\bm{u}^*(t) = -\bm{S}^{-1} \big[ \Ab^T \bm{g}(t) + \Ab^T \bm{H}(t) \bm{\lambda}(t) + \frac{1}{2} \diag(\Ab^T \bm{H}(t) \Ab) \big]
\end{align}
\end{theorem}

This optimal intensity is linear in $\lambdab(t)$ and the coefficients $\bm{g}(t)$ and $\bm{H}(t)$ can be found off-line. Hence,
it allows for a very efficient procedure to sample the times of the users'{} incentivized actions, which takes inspiration from the 
algorithm \redqueen~\cite{redqueen17wsdm}..
The key idea is as follows. 

At any given time $t$, we can view the multidimensional control signal $\bm{u}(t)$ as a superposition of
inhomogeneous multidimensional poisson processes, one per non incentivized action, which start when the actions take place.
Algorithm~\ref{alg:sampling} summarizes our method, which we name \cheshire~\cite{carroll1917through}. Within the algorithm,
\emph{NextAction}() returns the time of the next (non incentivized) action in the network as well as the identify of the user who 
takes the action, once the action happens, 
$\bm{e}_j$ is an indicator vector where the entry corresponding to user $j$ is $1$,
and $\emph{Sample}(\bm{u}(t))$ samples from a multidimensional inhomogeneous poisson process with intensity $\bm{u}(t)$ 
and it returns both the sampled time and dimension. 
To sample from a multidimensional inhomogeneous poisson process, there exist multiple methods \eg, refer to Lewis et al.~\cite{lewis1979simulation}.
Finally, note that one can precompute most of the quantities the algorithm needs, \ie, lines 2-3, $\Ab \bm{e}_j$ in line 8, and $\bm{S}^{-1} \Ab^T \bm{H}(t)$ 
in line 9. 
Given these precomputations, the algorithm only needs to perform $O(n)$ operations and sample $1^{T} \Nb(t_f)$ times from an inhomogeneous poisson 
process.

\section{Experiments}
\label{sec:experiments}
\begin{figure*}[t]
	\captionsetup[subfigure]{labelformat=empty}
	\centering
	\begin{tabular}{cccc}
	\subfloat[Uncontrolled, $\Nb(t_f)$]{ \includegraphics[width=0.20\textwidth]{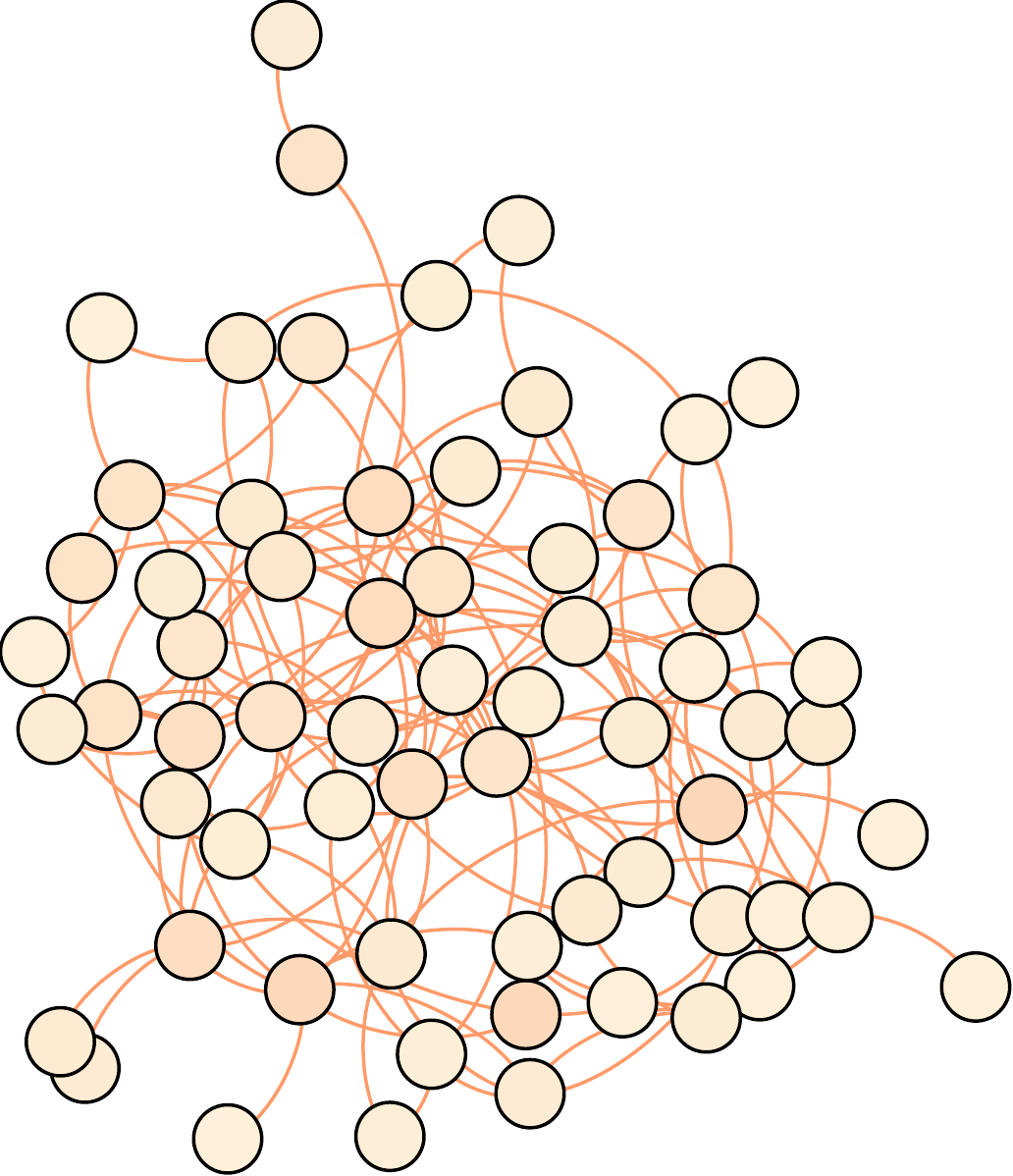} \label{fig:1_1}} &
	\subfloat[\cheshire, $\Nb(t_f)$]{ \includegraphics[width=0.20\textwidth]{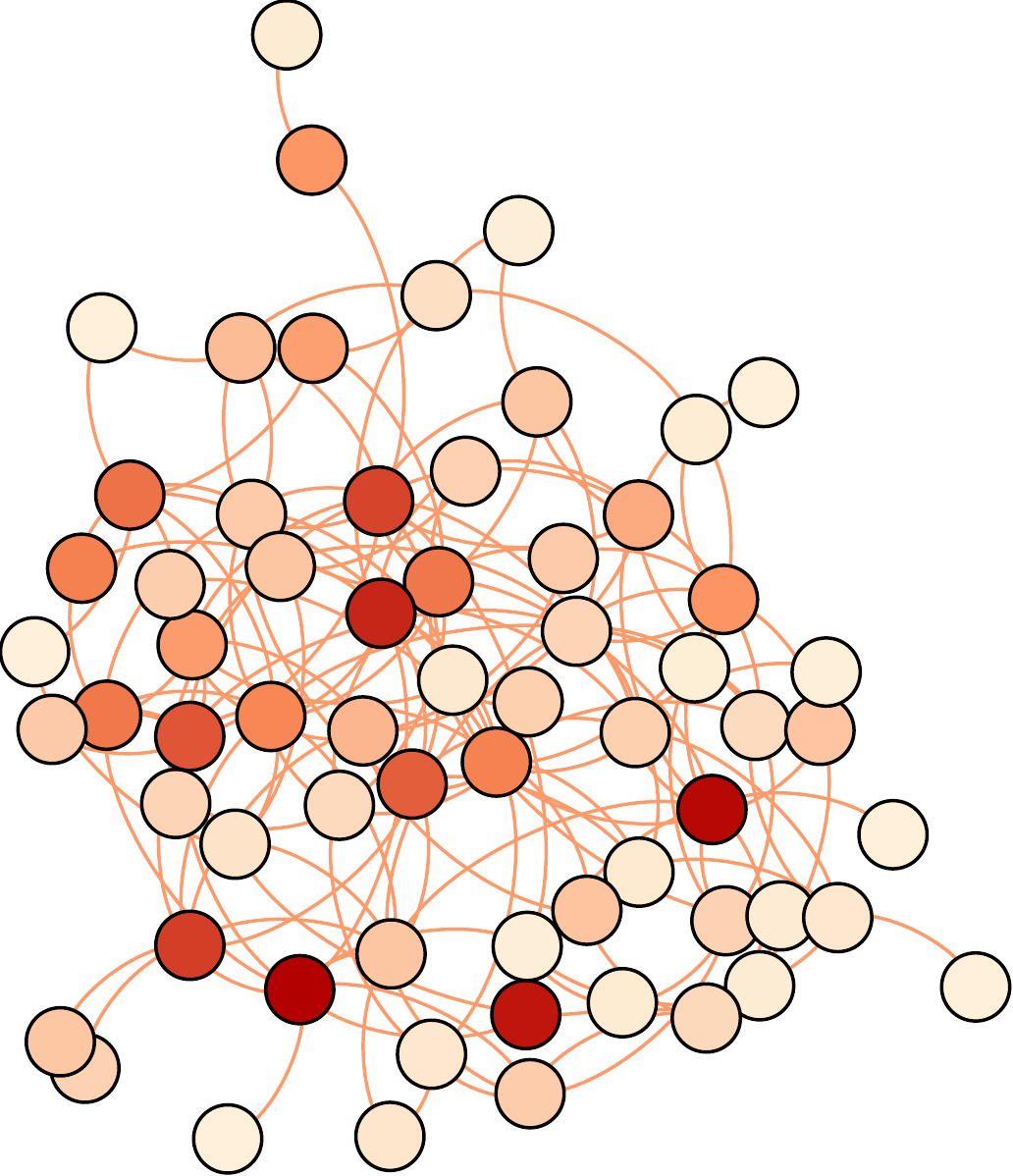} \label{fig:1_2}} &
	\subfloat[\cheshire, $\Mb(t_f)$]{ \includegraphics[width=0.20\textwidth]{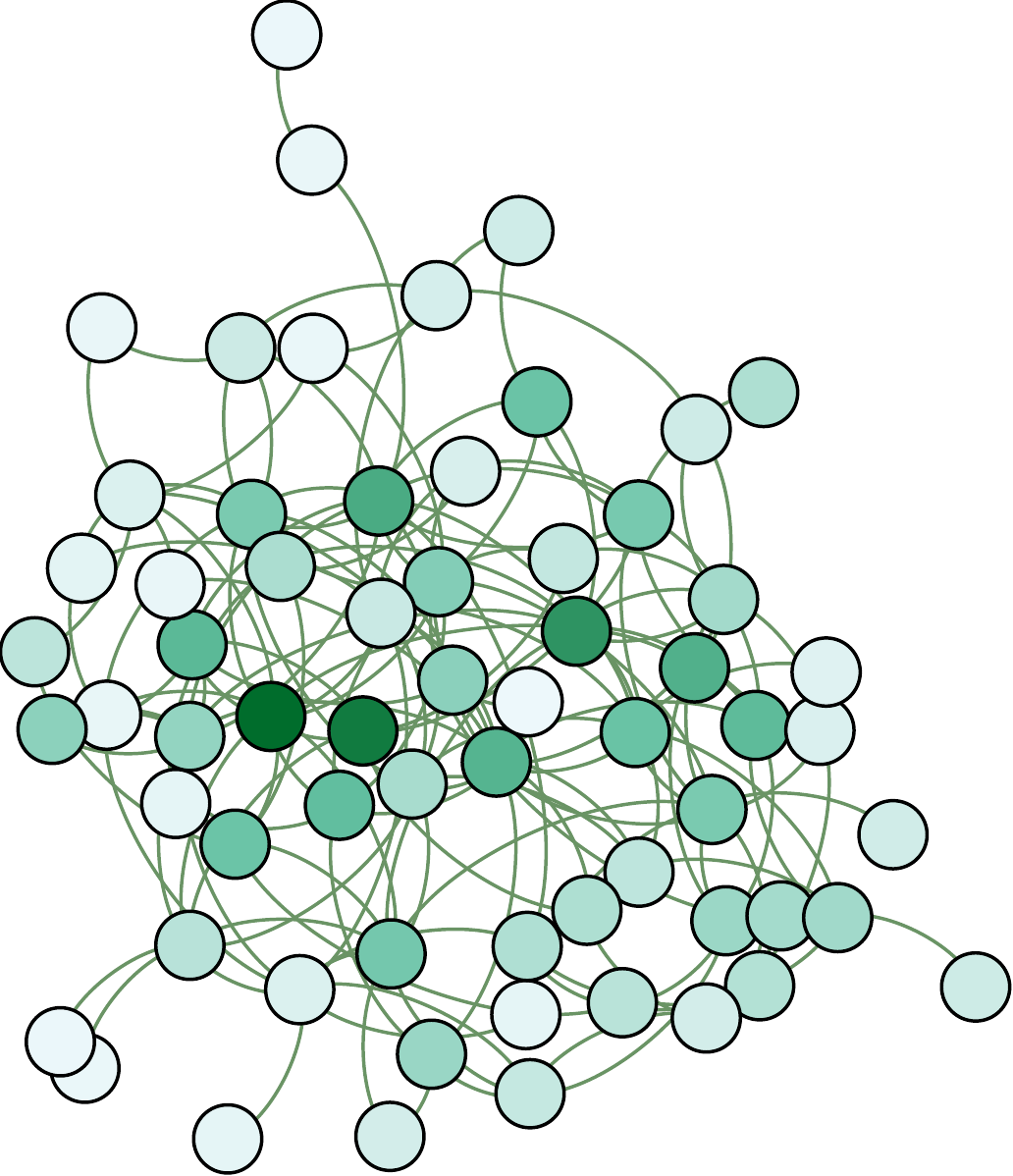} \label{fig:1_3}} &
 	\subfloat[Total number of actions]{\includegraphics[width=0.20\textwidth]{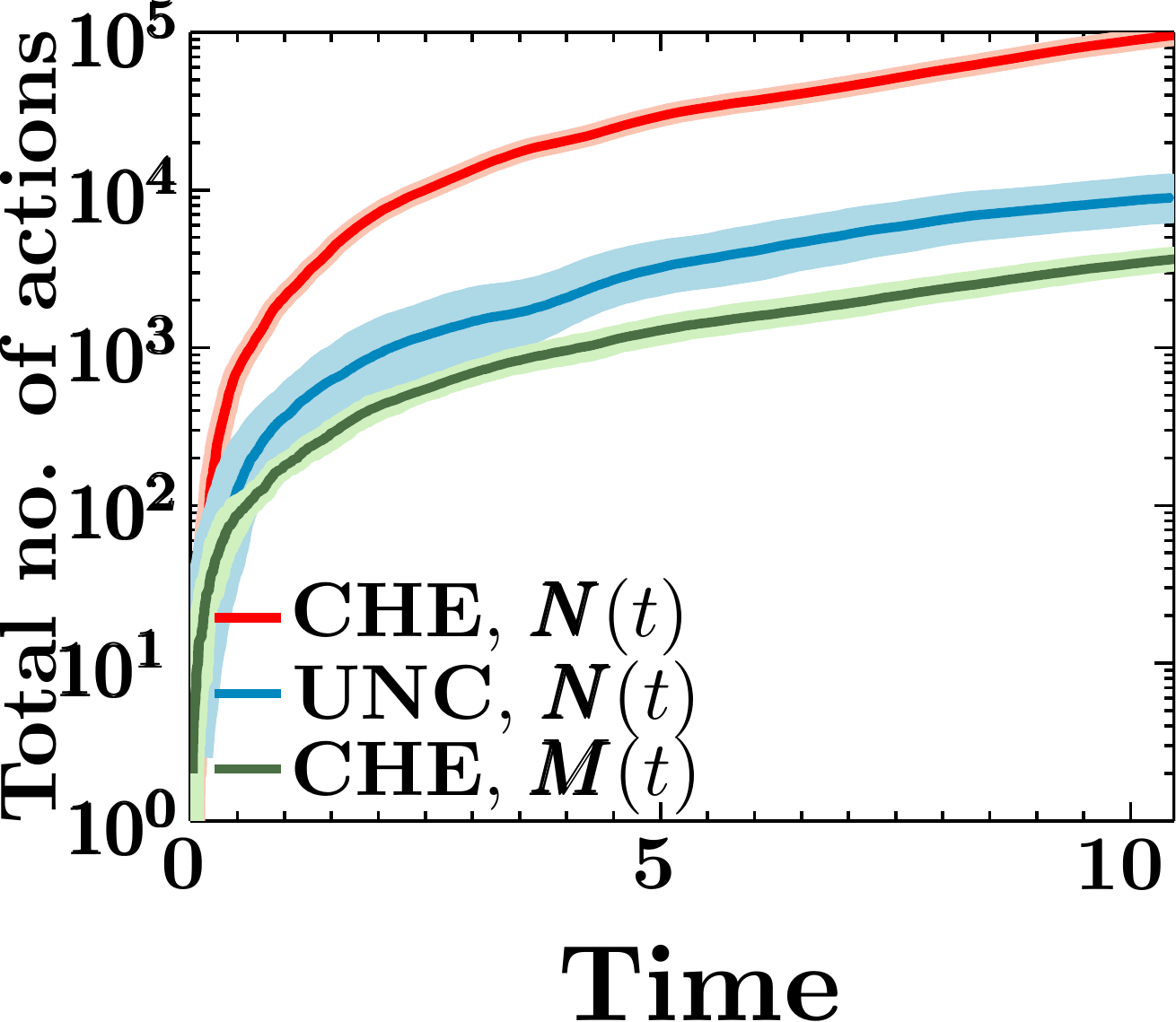} \label{fig:1_4}} \vspace{2mm} \\ 
         \multicolumn{4}{c}{} \hfill (a) Core-Periphery network \hfill  \vspace{1mm} \\
	\subfloat[Uncontrolled, $\Nb(t_f)$]{ \includegraphics[width=0.20\textwidth]{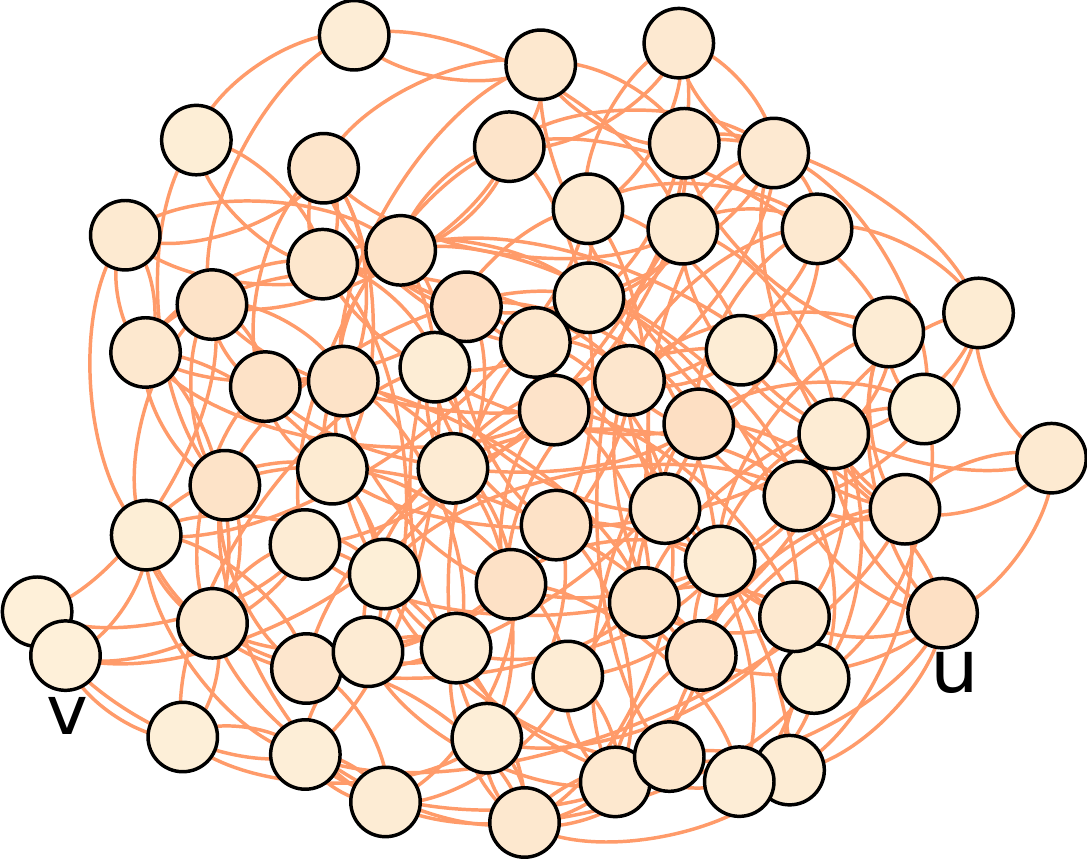} \label{fig:1_5}}  &
	\subfloat[\cheshire, $\Nb(t_f)$]{ \includegraphics[width=0.20\textwidth]{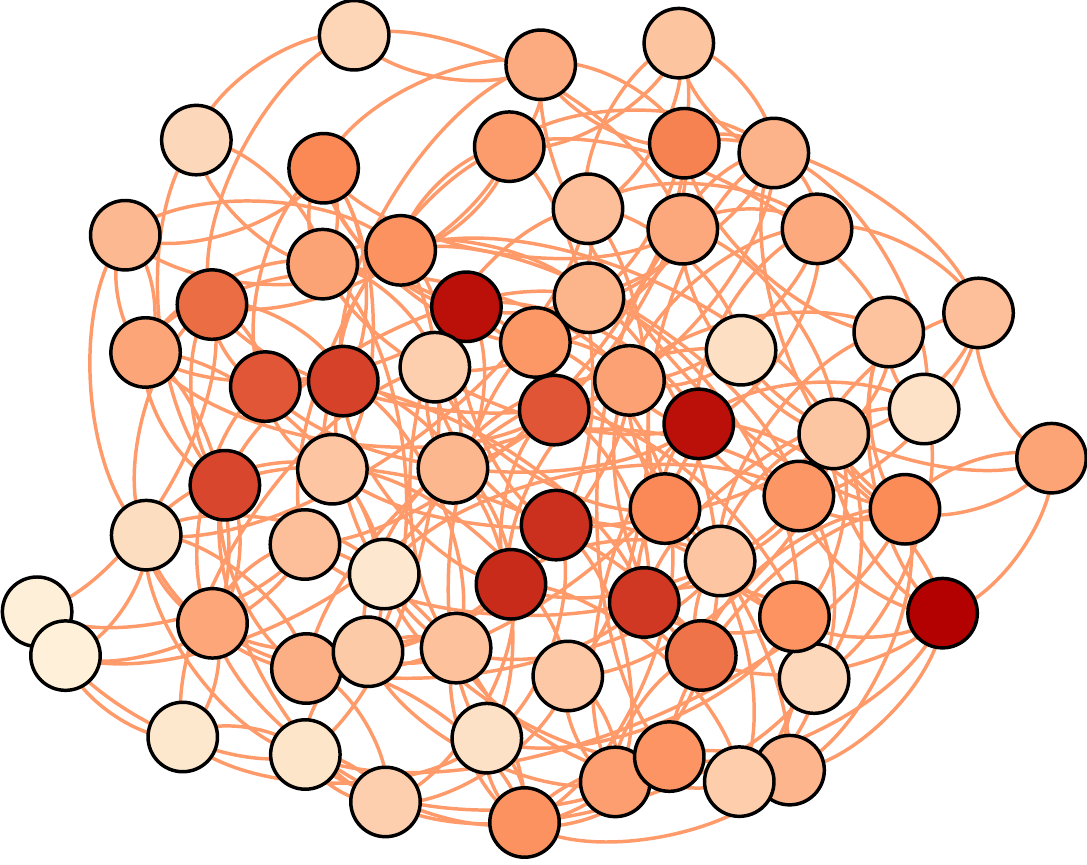} \label{fig:1_6}} &
	\subfloat[\cheshire, $\Mb(t_f)$]{ \includegraphics[width=0.20\textwidth]{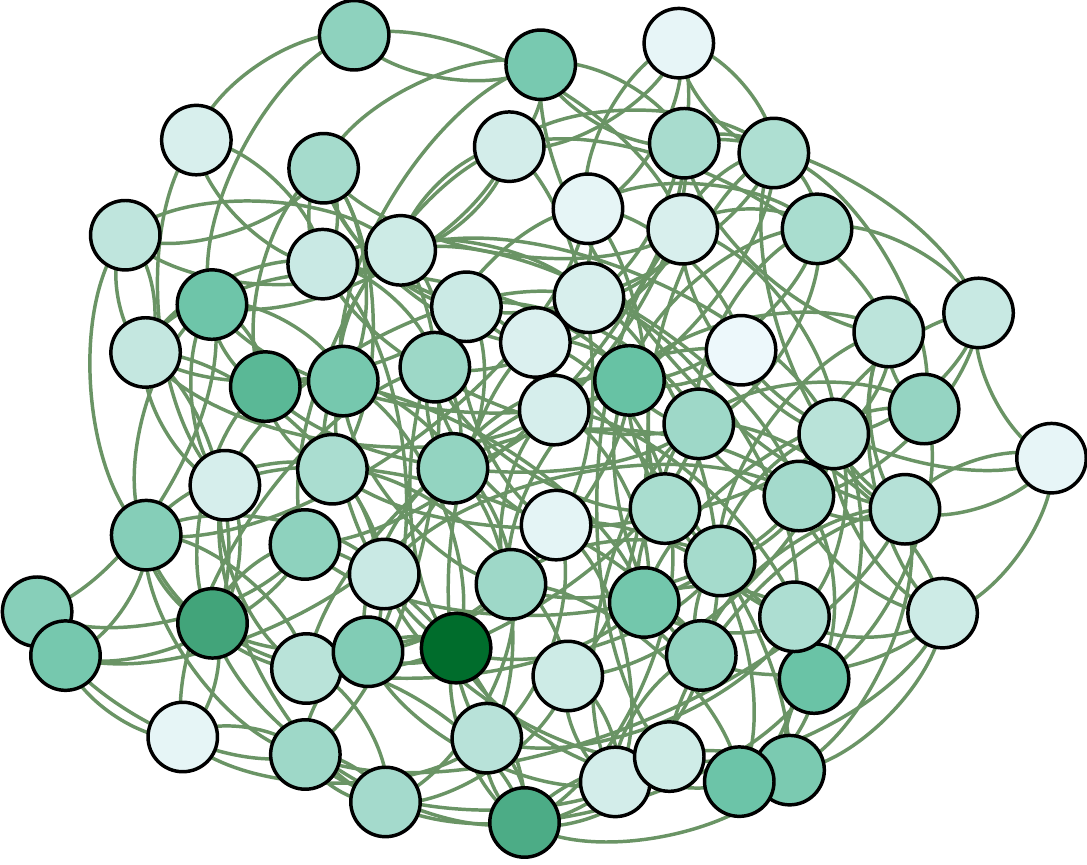} \label{fig:1_7}}  &
	\subfloat[Total number of actions]{\includegraphics[width=0.20\textwidth]{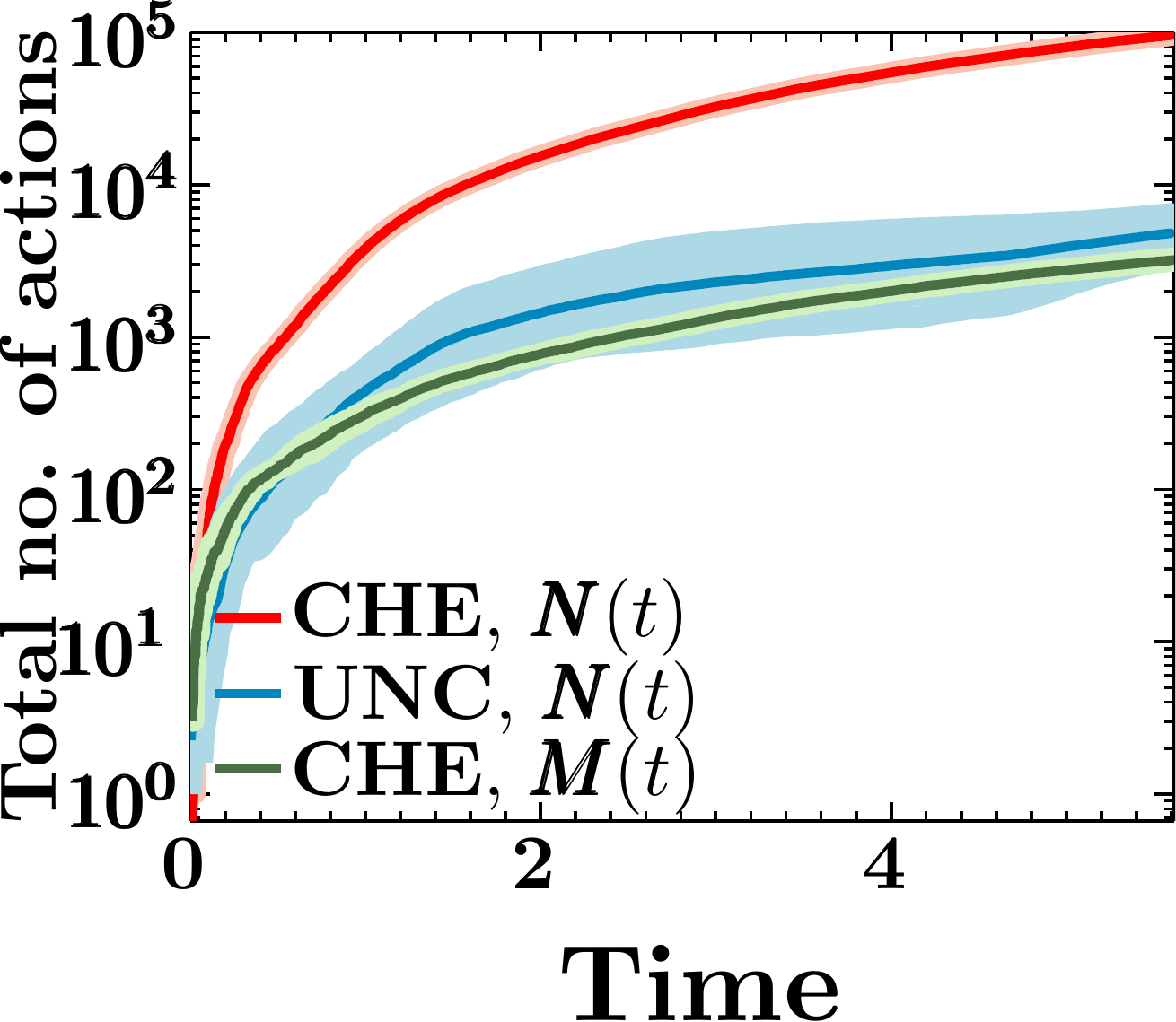} \label{fig:1_8}} \vspace{2mm} \\
	\multicolumn{4}{c}{} \hfill (b) Dissortative network \hfill \vspace{2mm} \\
	\end{tabular}
	        
 	\caption{Activity on two 64-node networks, $G_1$ (top) and $G_2$ (bottom) under uncontrolled (Eq.~\ref{eq:hawkes-dyn}; without \cheshire) and 
		      controlled (Eq.~\ref{eq:sys-dyn}; with \cheshire) dynamics.
		      The first and second columns visualize the final number of non incentivized actions $\bm{N}(t_f)$ under uncontrolled and controlled 
		      dynamics, where darker red corresponds to higher number of actions. 
		      The third column visualizes the final number of incentivized actions $\bm{M}(t_f)$ under controlled dynamics, where darker green corresponds
		      to higher number of actions.
		      The fourth column shows the temporal evolution of the number of incentivized and non incentivized actions across the whole networks for controlled 
		      and uncontrolled dynamics, where the solid line is the average across simulation runs and the shadowed region represents the standard error.
		      By incentivizing a relatively small number of actions ($\sim$3{,}600 actions), \cheshire is able to increase the overall number of (non incentivized) actions dramatically
		      ($\sim$$96{,}000$ vs $\sim$$4{,}800$ actions).
	}
	\label{fig:thdem}
	\vspace{-3mm}
\end{figure*}

\subsection{Experiments on synthetic data}
In this section, we shed light on \cheshire'{}s sampling strategy in two small Kronecker networks~\cite{Leskovec2010} by recording, on the one hand, the number of incentivized actions per node 
and, on the other hand, the number of additional actions per node these incentivized actions triggers in comparison with an uncontrolled setup. 
Additionally, Appendix~\ref{app:synthetic} compares the performance of our method against several baselines and state of the art methods~\cite{Farajtabar2015, farajtabar2016msc} on a variety of large synthetic 
networks and provides a scalability analysis.

\xhdr{Experimental setup}
We experiment with two small Kronecker networks with 64 nodes, a small core-periphery (parameter matrix $[0.96 , 0.3; 0.3, 0.96]$) and a dissortative network ($[0.3, 0.96 ; 0.96,0.3]$), shown 
in Figure~\ref{fig:thdem}.
For each network, we draw $\Ab$ from a uniform distribution $U(0,10)$, $\mub$ also from a uniform distribution $U(0, 10)$ for 
$20$\% of the nodes and $\mub = 0$ for the remaining $80$\%, and set $\omega=16$, $t_0 = 0$ and $t_f = 5.5$.
Then, we compare the number of actions over time under uncontrolled dynamics (Eq.~\ref{eq:hawkes-dyn}; without \cheshire) 
and controlled dynamics (Eq.~\ref{eq:sys-dyn}; with \cheshire). 
In both cases, we perform $20$ simulation runs and sample (non incentivized) actions from a multidimensional Hawkes process using Ogata'{}s thinning algorithm~\cite{Ogata1981}.
In the case of controlled dynamics, we sample incentivized actions using Algorithm~\ref{alg:sampling}.

\xhdr{Results} Figure~\ref{fig:thdem} summarizes the results in terms of the number of non incentivized and incentivized actions, which show that: 
(i) a relatively small number of incentivized actions (fourth column; $\sim$3{,}600 actions) results in a dramatic increase in the overall number of (non incentivized) actions with respect the uncontrolled 
setup (fourth column; $\sim$$96{,}000$ vs $\sim$$4{,}800$ actions); 
(ii) the majority of the incentivized actions concentrate in a small set of influential nodes (third column; nodes in dark green);  
and, (iii) the variance of the overall number of actions (fourth column; shadowed regions) is consistently reduced when 
using \cheshire, in other words, the networks become more robust.

\subsection{Experiments on real data}
In this section, we experiment with data gathered from Twitter and show that our model can maximize the number of online actions 
more effectively than several baselines and state of the art methods~\cite{Farajtabar2015, farajtabar2016msc}.

\xhdr{Experimental setup}
We experiment with five Twitter datasets about current real-world events (Elections, Verdict, Club, Sports, TV Show), where actions are
tweets (and retweets).
Appendix~\ref{app:real-datasets} contains further details and statistics about these datasets.
We compare the performance of our algorithm, \cheshire, with two state of the art methods, which we denote as OPL~\cite{Farajtabar2015} and 
MSC~\cite{farajtabar2016msc}, and two baselines, which we denote as PRK and DEG. 
PRK and DEG distribute the users'{} control intensities $\mathbf{u}(t)$ proportionally to the user'{}s page rank and outgoing degree in the network, 
respectively.
More in detail, we proceed as follows.
\begin{figure*}[t]
\captionsetup[subfigure]{labelformat=empty}
\centering
  { \includegraphics[width=0.956\textwidth]{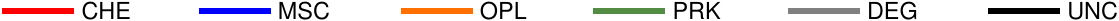}}\\ \vspace*{-2mm}
 \subfloat[\small Elections, $\bar{M}(t_f)\approx 3K$]{\setcounter{subfigure}{1} \includegraphics[height=0.16\textwidth]{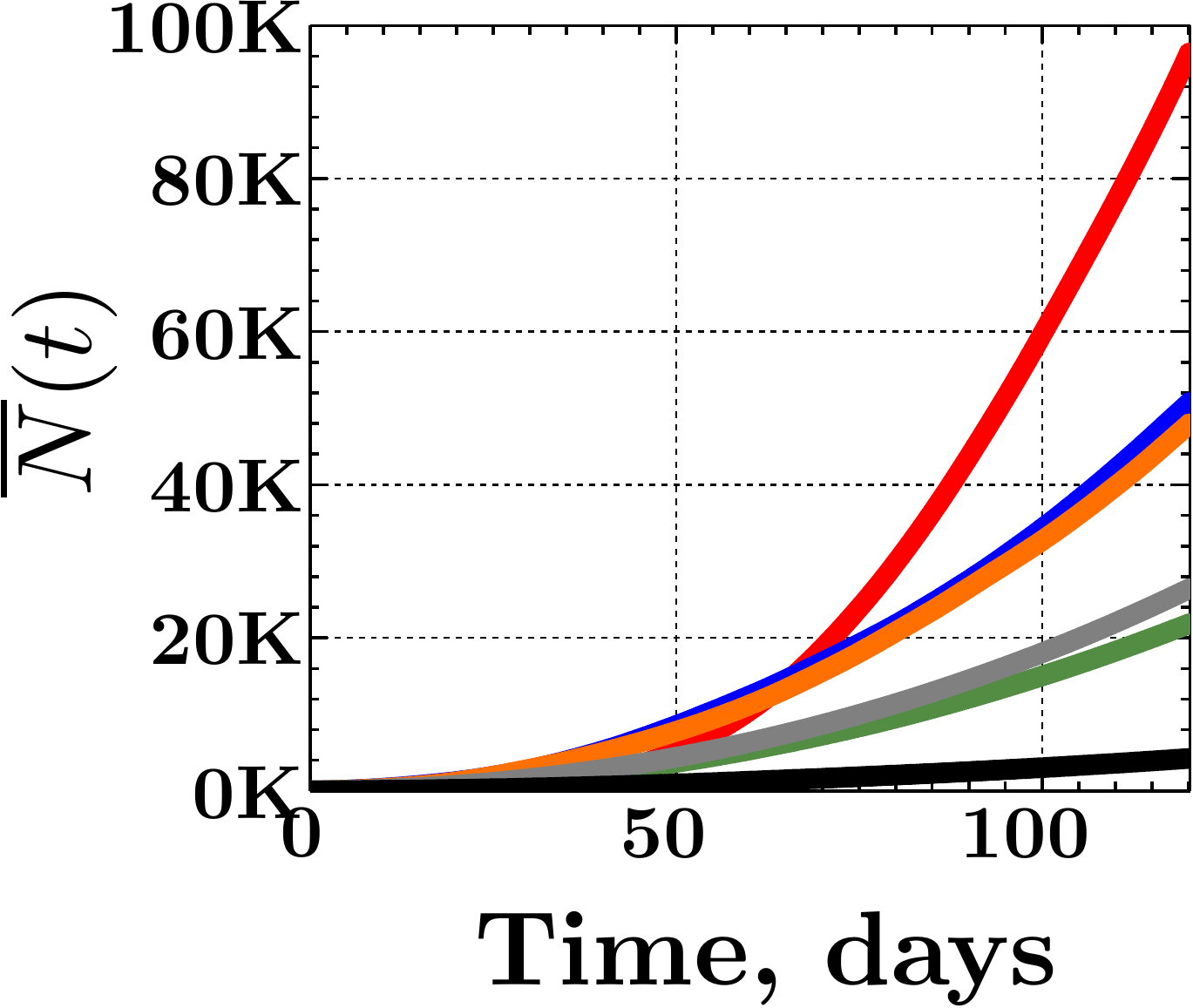}}
\hspace*{2mm}\subfloat[\small  Verdict, $\bar{M}(t_f)$$\approx$$14K$]{\includegraphics[height=0.16\textwidth]{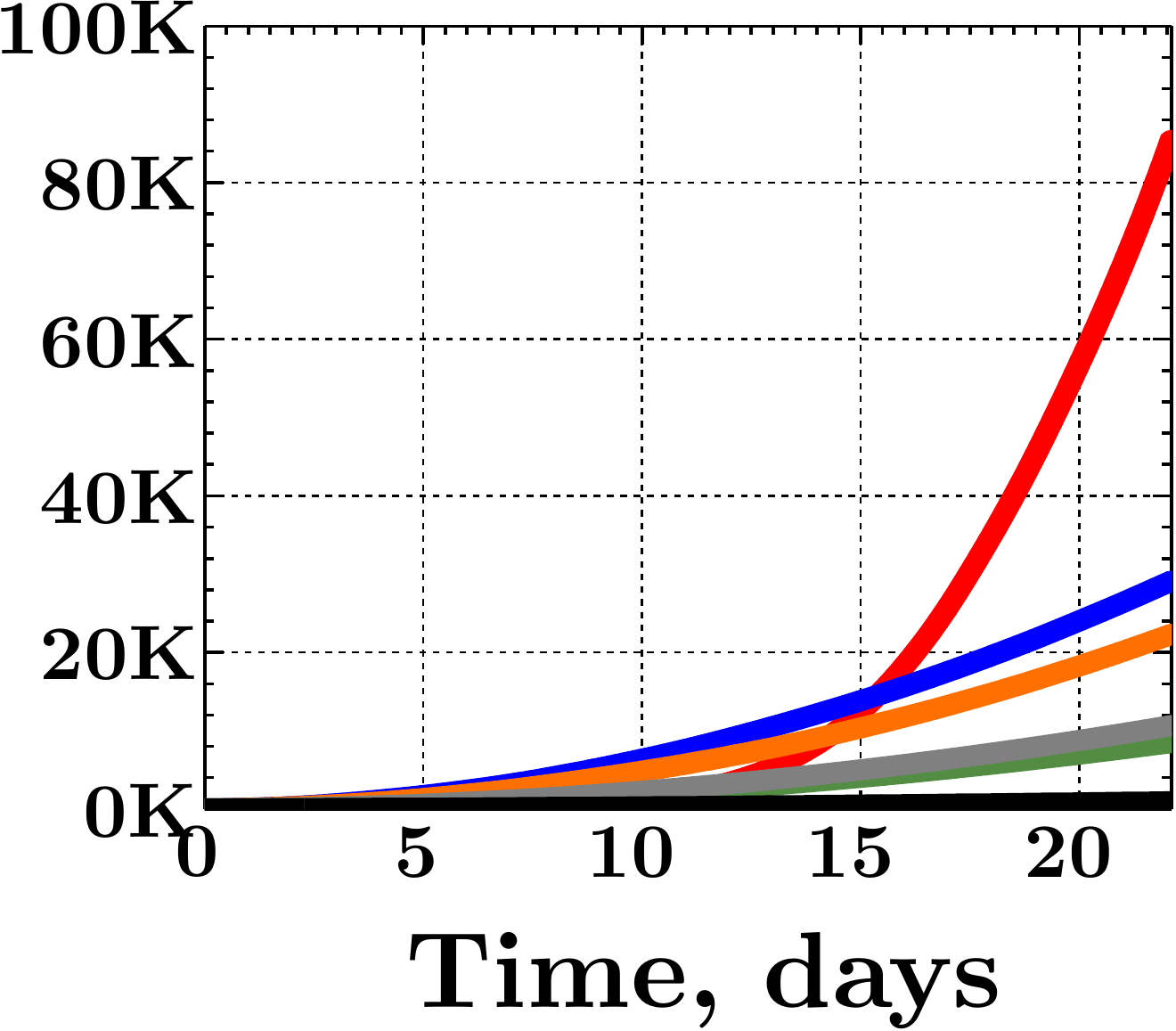}}
\hspace*{2mm}\subfloat[\small  Club, $\bar{M}(t_f)\approx 5K$ ]{\includegraphics[height=0.16\textwidth]{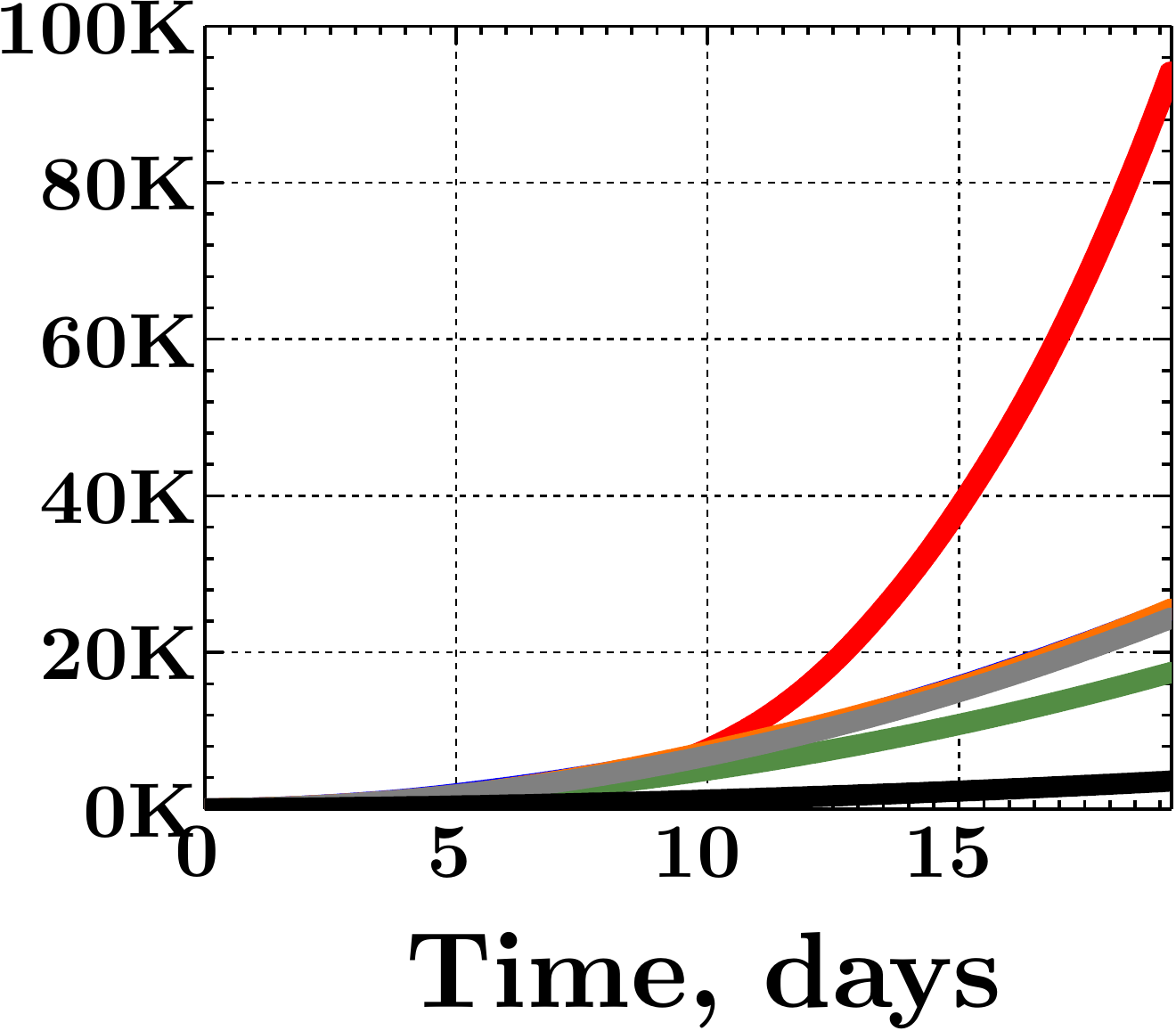}}  
\hspace*{2mm}\subfloat[\small  Sports, $\bar{M}(t_f)\approx 5K$]{ \includegraphics[height=0.16\textwidth]{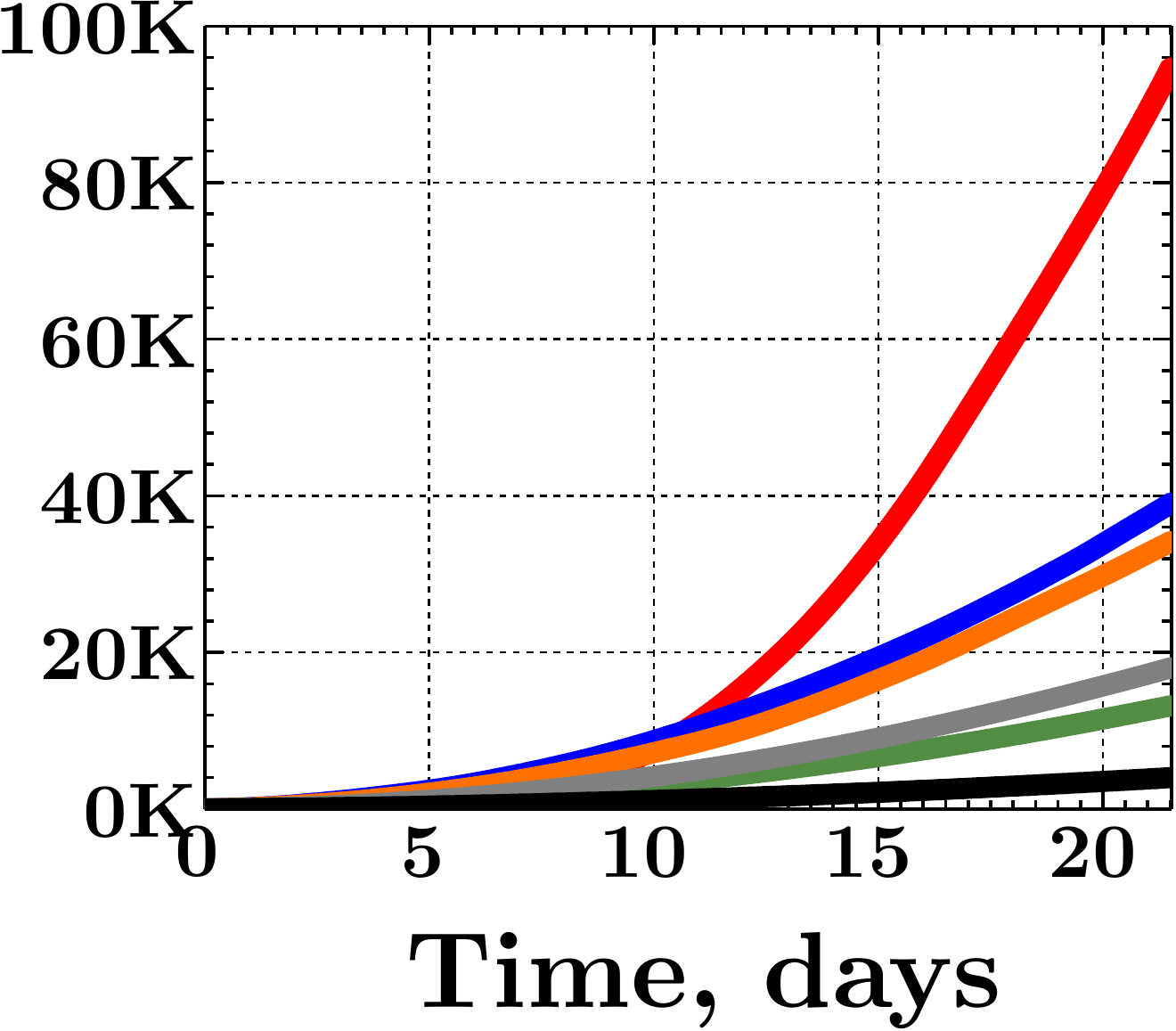}}
\hspace*{2mm}\subfloat[\small  Series, $\bar{M}(t_f)\approx 13K$]{ \includegraphics[height=0.16\textwidth]{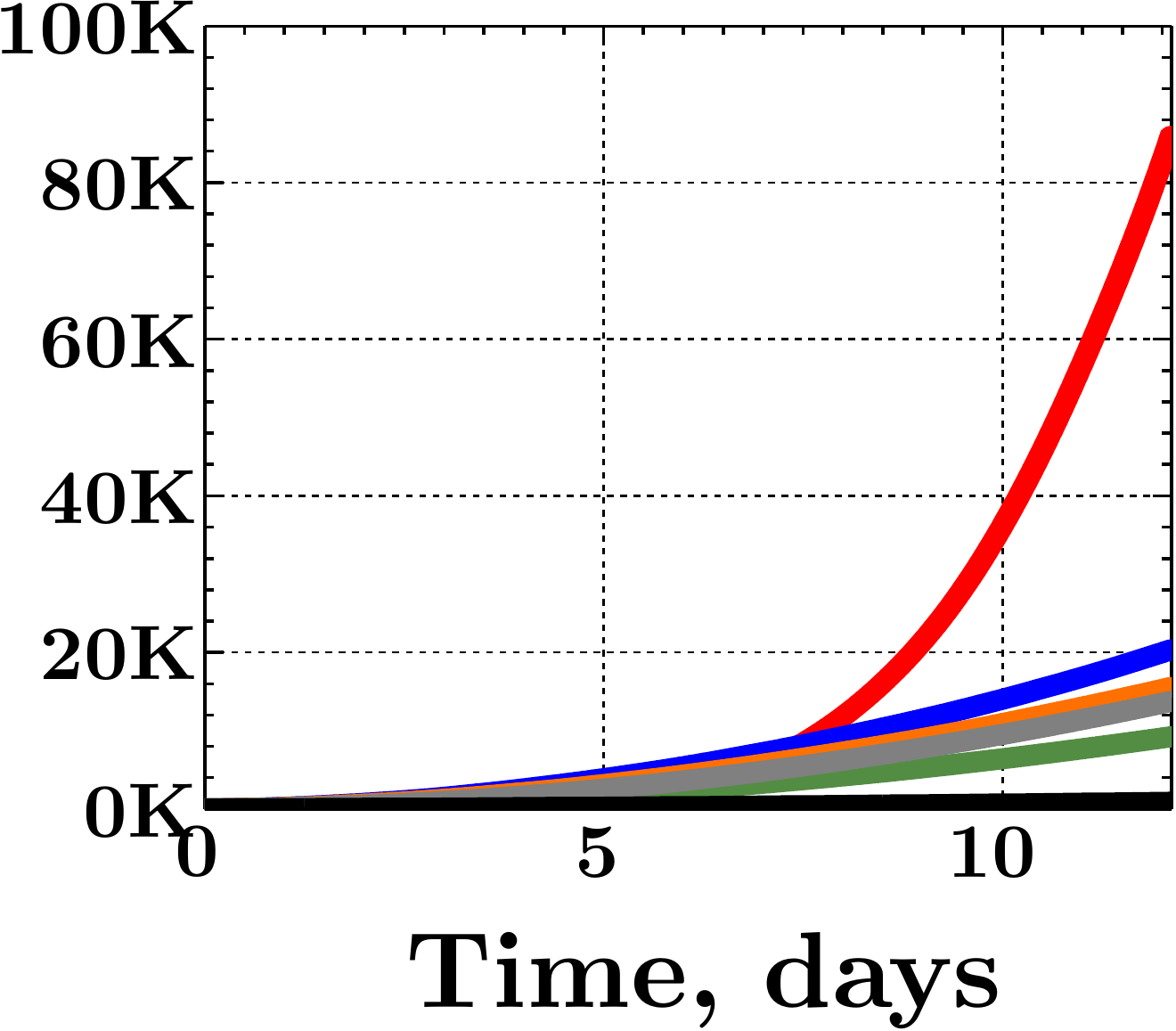}}

\caption{Performance over time of \cheshire against several competitors for each Twitter dataset. Performance is measured in terms of overall number of tweets 
$\bar{N}(t) = \sum_{u \in \Vcal} \EE[N_u(t)]$. In all cases, we tune the parameters $\bm{Q}$, $\bm{S}$ and $\bm{F}$ to be diagonal
matrices such that the total number of incentivized tweets \emph{posted} by our method is equal to the budget used in the competing methods and baselines.
\cheshire (in red) consistently outperforms competing methods over time and it triggers up to $100$\%--$400$\% more posts than the second best performer (in blue) as time 
goes by.}
\label{fig:NwT}
\end{figure*}

For each dataset, we estimate the influence matrix $\Ab$ of the multidimensional Hawkes process defined by Eq.~\ref{eq:multi-hawkes} using maximum likelihood, 
as elsewhere~\cite{Farajtabar2014,Valera2015}. Moreover, we set the decay parameter $\omega$ of the corresponding exponential triggering kernel $\kappa(t)$ by 
cross-validation.
Then, we perform $20$ simulation runs for each method and baseline, where we sample non incentivized actions from the multidimensional Hawkes process learned
from the corresponding Twitter dataset using Ogata'{}s thinning algorithm~\cite{Ogata1981}. 
For the competing methods and baselines, the control intensities $\bm{u}(t)$ are deterministic and
thus we only need to sample incentivized actions from inhomogeneous poisson processes~\cite{lewis1979simulation}. 
For our method, the control intensities are stochastic and thus we sample incentivized actions using Algorithm~\ref{alg:sampling}.
Finally, we compare their performance in terms of the (empirical) average number of tweets $\EE[\Nb(t)]$.
%
In the above procedure, for a fair comparison, we tune the parameters $\bm{Q}$, $\bm{S}$ and $\bm{F}$ to be diagonal matrices such that the total number of incentivized tweets \emph{posted} by our method 
is equal to the budget used in the state of the art methods and baselines.

\xhdr{Results}
We first compare the performance of our algorithm against others in terms of overall average number of tweets $\bar{N}(t) = \sum_{u \in \Vcal} \EE[N_u(t)]$ for a fixed budget 
$\bar{M}(t_f) = \sum_{u \in \Vcal} \EE[M_u(t_f)]$. Figure~\ref{fig:NwT} summarizes the results, which show that: (i) our algorithm consistently outperforms competing methods
by large margins; (ii) it triggers up to $100$\%--$400$\% more posts than the second best performer as time goes by; and, (iii) the baselines PRK and DEG have 
an underwhelming performance, suggesting that the network structure alone is not an accurate measure of influence.

\begin{figure*}[t]
\captionsetup[subfigure]{labelformat=empty}
\centering
 { \includegraphics[width=0.956\textwidth]{legend_IwB_new}}\\ \vspace*{-2mm}
 \hspace*{-0.6cm}\subfloat[Elections]{\setcounter{subfigure}{1} \includegraphics[height=0.18\textwidth]{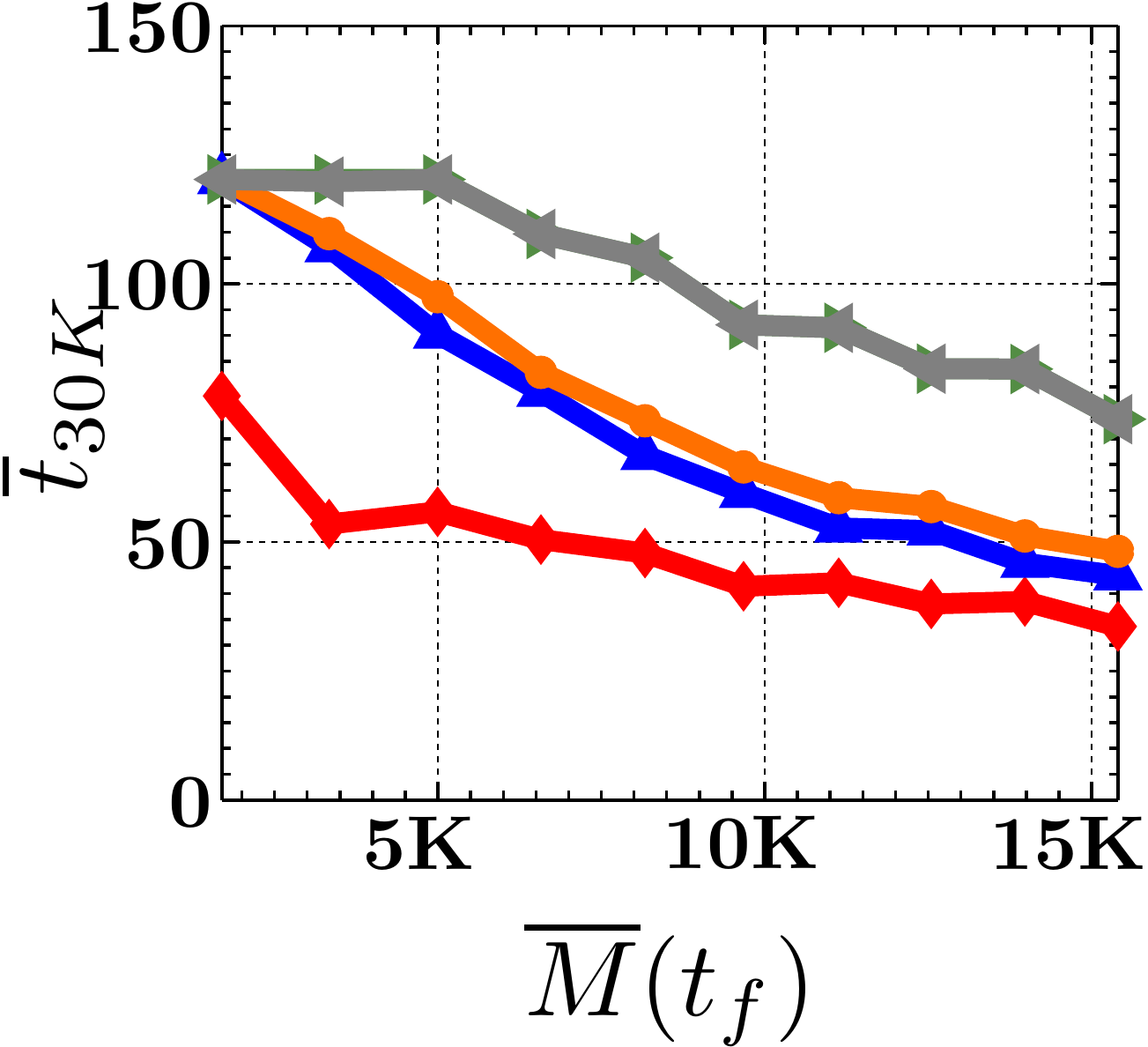}}
\hspace*{0.1cm}\subfloat[Verdict]{\includegraphics[height=0.18\textwidth]{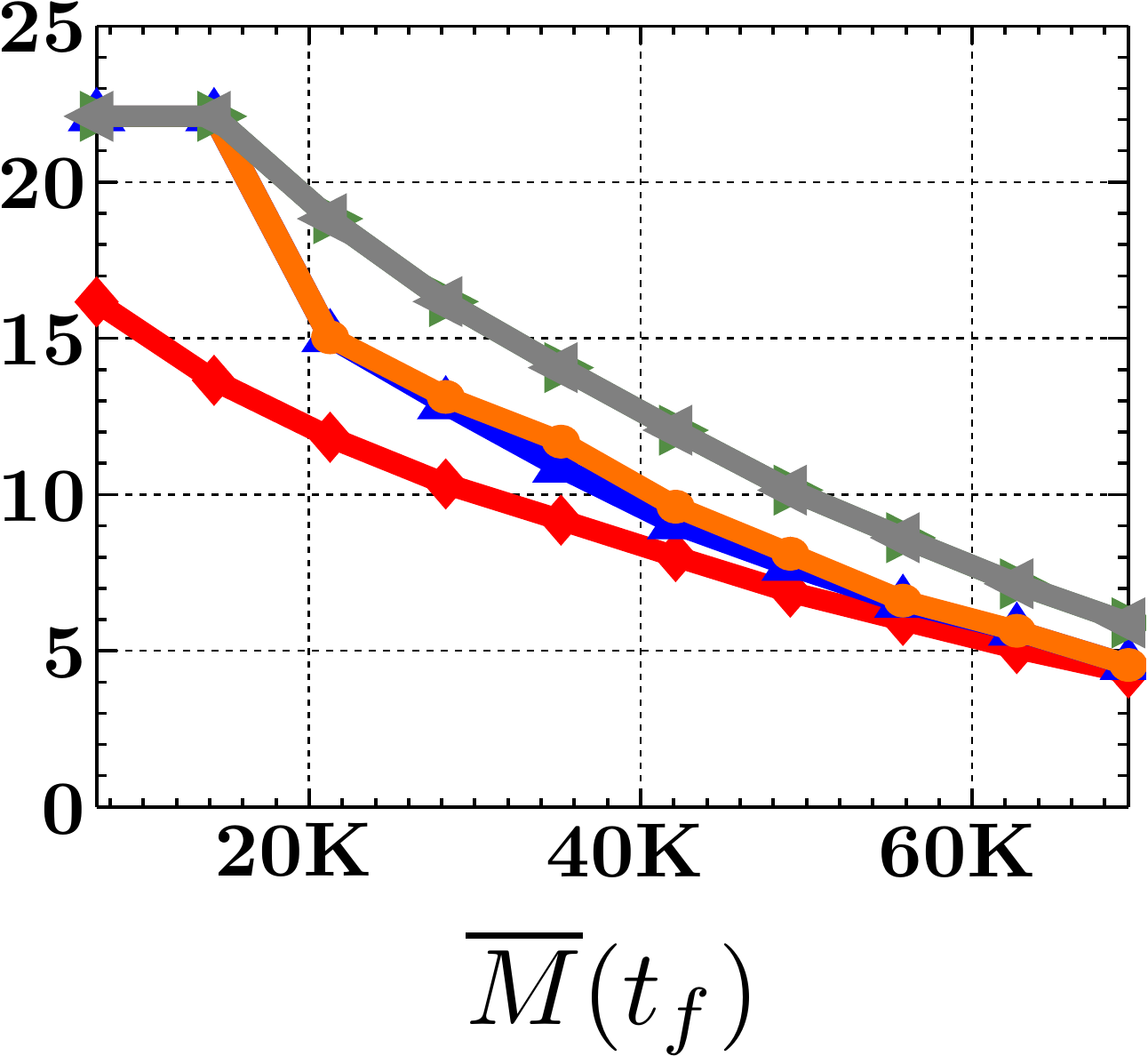}}
\hspace*{0.1cm}\subfloat[Club]{\includegraphics[height=0.18\textwidth]{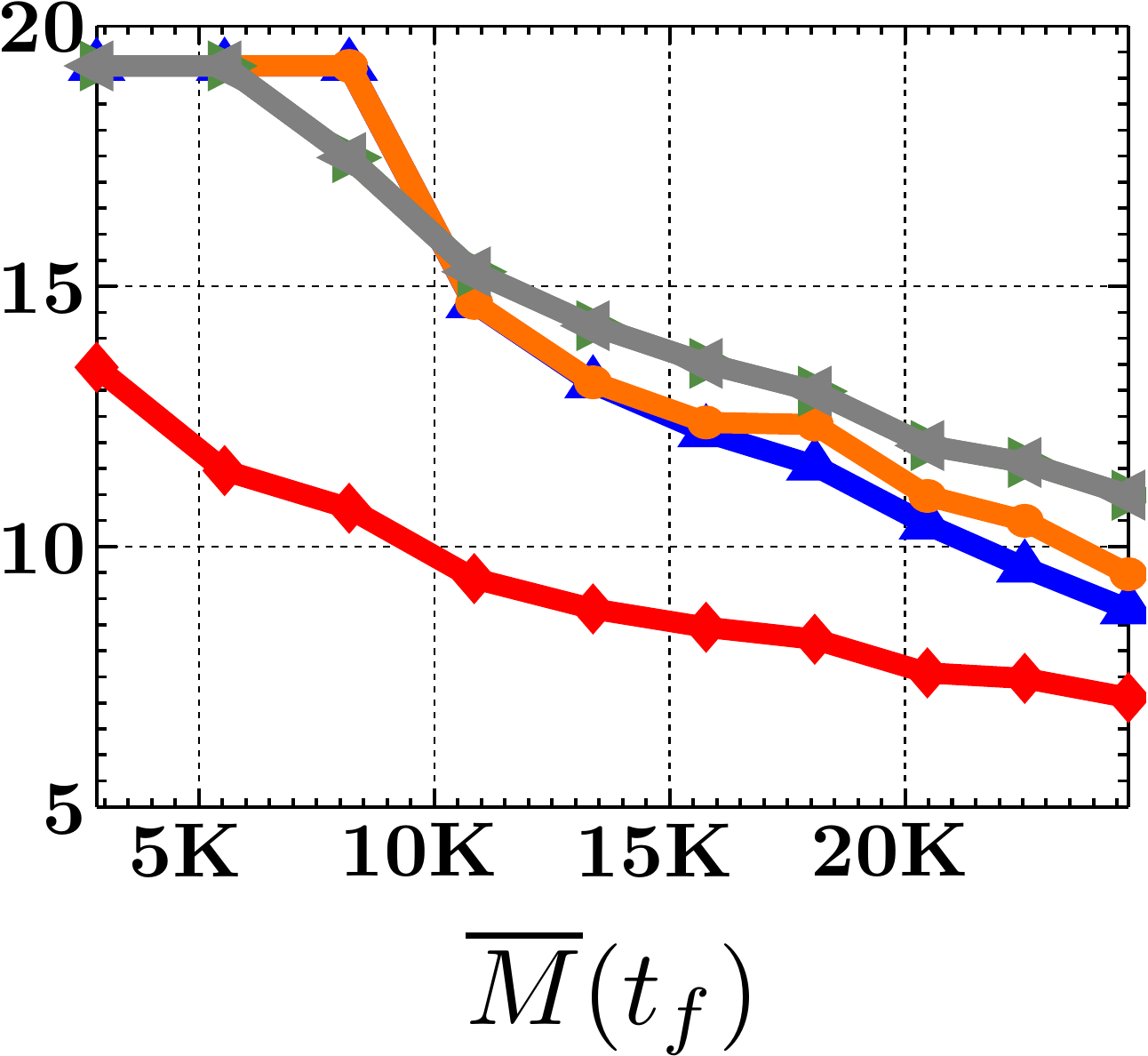}}  
\hspace*{0.1cm}\subfloat[Sports]{ \includegraphics[height=0.18\textwidth]{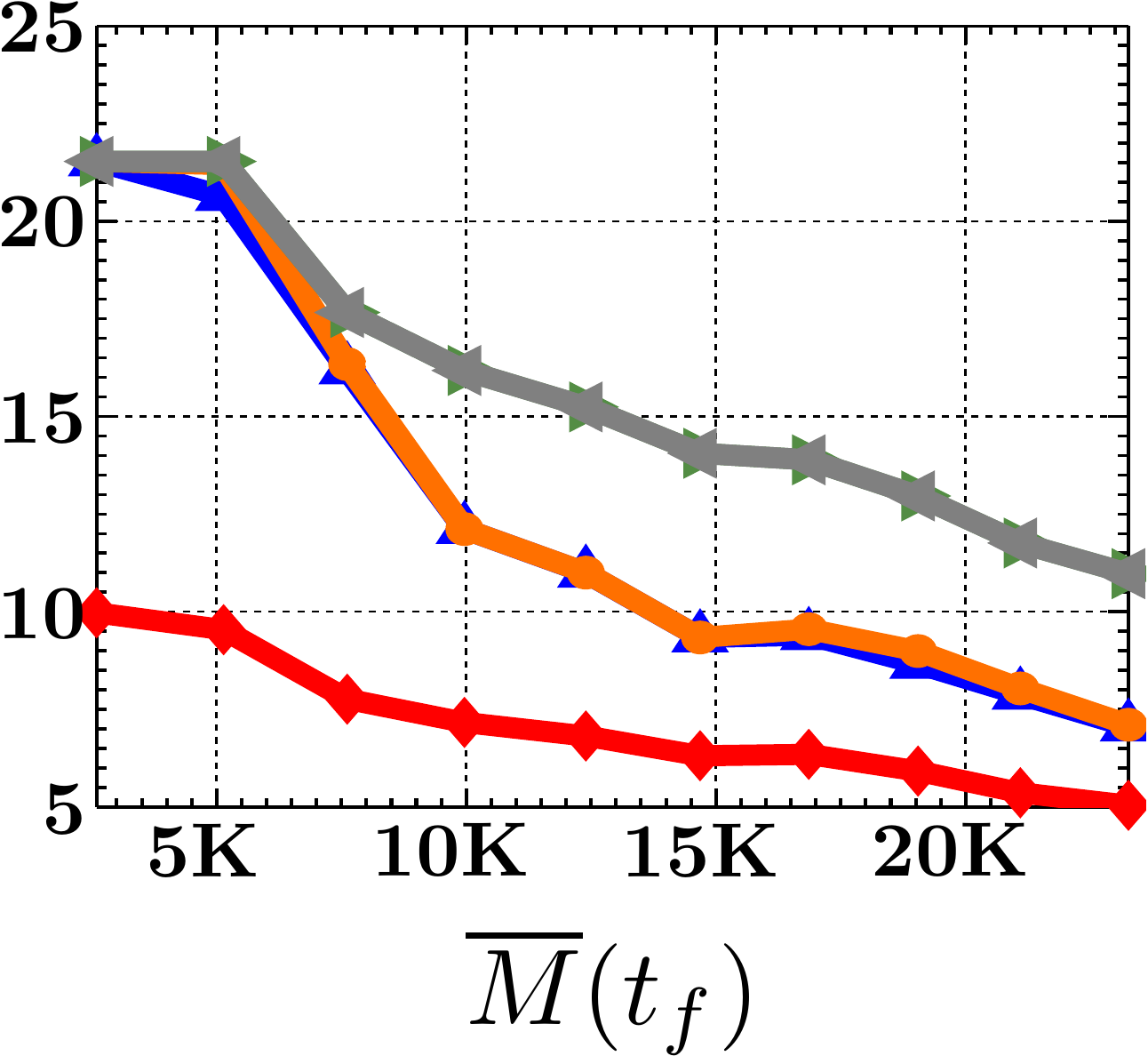}}
 \hspace*{0.1cm}\subfloat[Series]{ \includegraphics[height=0.18\textwidth]{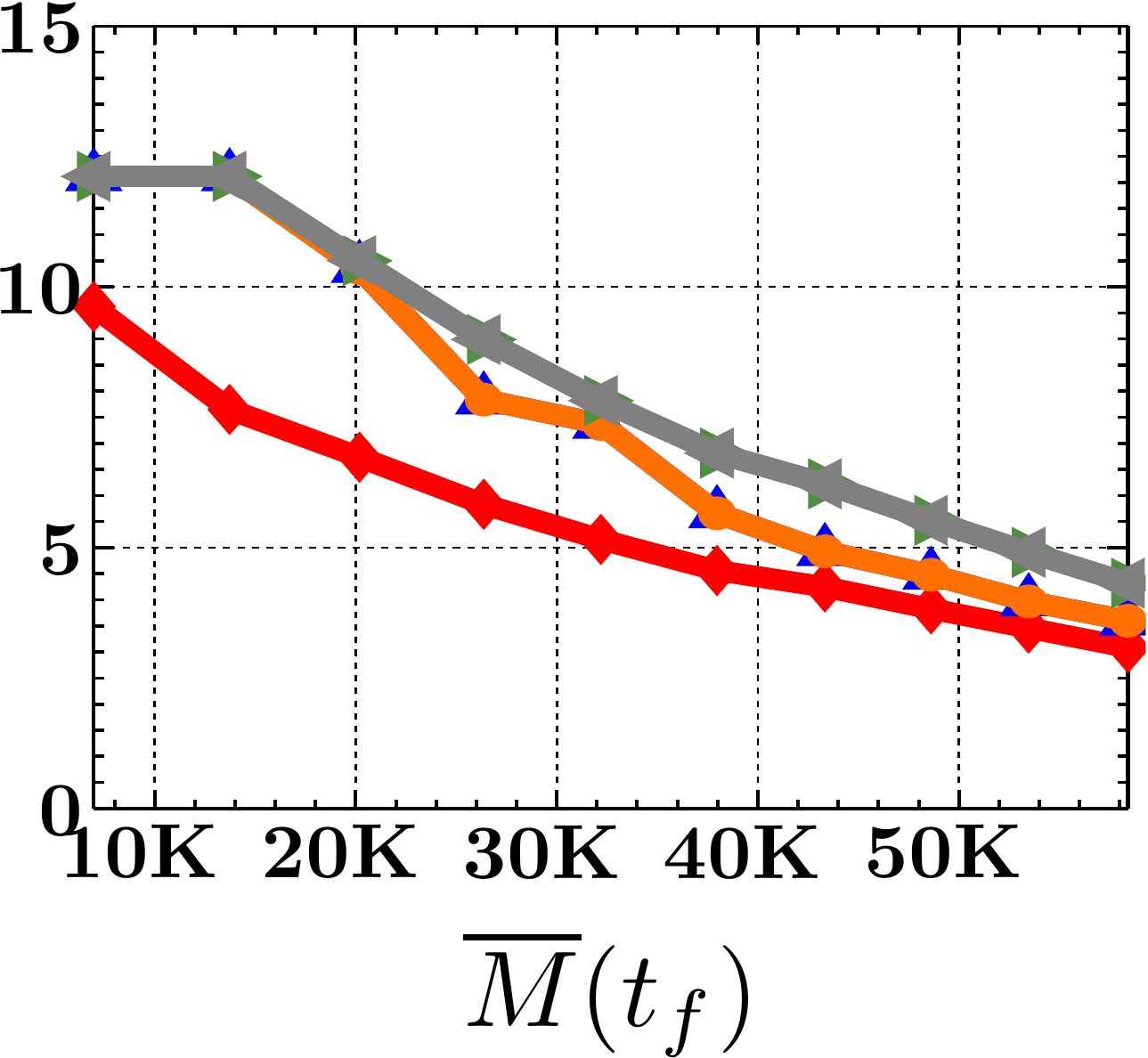}}

\caption{Performance vs. number of incentivized tweets for each Twitter dataset. Performance is measured in terms of the average time $\bar{t}_{30K}$ required by each 
method to reach a milestone of $30{,}000$ tweets.
\cheshire (in red) consistently reaches the milestone faster than the competing methods, \eg, $20$\%--$50$\% faster than the second best performer (in blue) for low
budgets.}
%
\label{fig:NuT}
\end{figure*}

Next, we evaluate the performance of our algorithm against others with respect to the available budget. To this aim, 
we compute the average time $\bar{t}_{30K}$ required by each method to reach a milestone of $30{,}000$ tweets against the number of incentivized tweets 
$\bar{M}(t_f)$ (\ie, the budget). Here, we do not report the results for the uncontrolled dynamics (UNC) since it did not reach the milestone after $10$$\times$ the time the 
slowest competitor took to reach it.
Figure~\ref{fig:NuT} summarizes the results, which show that: (i) our algorithm consistently reaches the milestone faster than the competing 
methods; (ii) it exhibits a greater competitive advantage when the budget is low; and, (iii) it reaches the milestone $20$\%--$50$\% faster than the second best performer
for low budgets.

\section{Conclusions}
\label{sec:conclusions}
In this paper, we tackle the problem of activity maximization in social networks from the perspective of stochastic optimal control of temporal point
processes and showed that the optimal level of incentivized actions depends linearly on the current level of overall actions. Moreover, the coefficients 
of this linear relationship can be found by solving a matrix Riccati differential equation, which can be solved efficiently, and a first order differential 
equation, which has a closed form solution. 
As a result, we were able to design \cheshire (Algorithm~\ref{alg:sampling}), an efficient and relatively simple online algorithm to sample the optimal 
times of the users'{} incentivized actions.
We experimented with synthetic and real-world data gathered from Twitter and showed that our algorithm is able to consistently maximize activity more 
effectively than the state of the art.

Our work also opens many interesting venues for future work. 
For example, \cheshire optimizes a sum of quadratic losses on the users'{} intensities and assumes the influence matrix does not change over time, however, it would 
be useful to derive the optimal level of incentivized actions for other losses, \eg, minimax loss, and consider time-varying influence.
One of the challenges one would face is to find solutions that ensure the nonnegativity of the control intensity.
Moreover, our experimental evaluation is based on simulation, using models whose parameters ($\Ab$, $\omega$) are learned from data. It would be very interesting to 
evaluate our method using actual interventions in a social network.
Finally, optimal control of jump SDEs with doubly stochastic temporal point processes can be potentially applied to design online algorithms for a wide variety of control problems 
in social and information systems, such as human learning~\cite{reddy2016unbounded}, opinion control~\cite{wang2017}, and rumor control~\cite{friggeri2014rumor}.

\bibliographystyle{abbrv}
\bibliography{refs}

\begin{thebibliography}{10}

\bibitem{Aalen2008}
O.~Aalen, O.~Borgan, and H.~Gjessing.
\newblock {\em Survival and event history analysis: a process point of view}.
\newblock Springer Science \& Business Media, 2008.

\bibitem{bertsekas1995dynamic}
D.~Bertsekas.
\newblock {\em Dynamic programming and optimal control}, volume~1.
\newblock Athena Scientific Belmont, MA, 1995.

\bibitem{carroll1917through}
L.~Carroll.
\newblock {\em Through the looking glass: And what Alice found there}.
\newblock Rand, McNally, 1917.

\bibitem{CheWanWan2010}
W.~Chen, C.~Wang, and Y.~Wang.
\newblock Scalable influence maximization for prevalent viral marketing in
  large-scale social networks.
\newblock In {\em Proceedings of the 16th ACM SIGKDD international conference
  on Knowledge discovery and data mining}, pages 1029--1038. ACM, 2010.

\bibitem{de2016learning}
A.~De, I.~Valera, N.~Ganguly, S.~Bhattacharya, and M.~Gomez-Rodriguez.
\newblock Learning and forecasting opinion dynamics in social networks.
\newblock In {\em Advances in Neural Information Processing Systems}, 2016.

\bibitem{Du2013}
N.~Du, L.~Song, M.~Gomez-Rodriguez, and H.~Zha.
\newblock Scalable influence estimation in continuous-time diffusion networks.
\newblock In {\em Advances in Neural Information Processing Systems}, pages
  3147--3155, 2013.

\bibitem{Farajtabar2014}
M.~Farajtabar, N.~Du, M.~Gomez-Rodriguez, I.~Valera, H.~Zha, and L.~Song.
\newblock Shaping social activity by incentivizing users.
\newblock In {\em Advances in neural information processing systems}, pages
  2474--2482, 2014.

\bibitem{Farajtabar2015}
M.~Farajtabar, M.~Gomez-Rodriguez, N.~Du, M.~Zamani, H.~Zha, and L.~Song.
\newblock Back to the past: Source identification in diffusion networks from
  partially observed cascades.
\newblock In {\em Proceedings of the 18th International Conference on
  Artificial Intelligence and Statistics}, 2015.

\bibitem{farajtabar2015coevolve}
M.~Farajtabar, Y.~Wang, M.~Gomez-Rodriguez, S.~Li, H.~Zha, and L.~Song.
\newblock {COEVOLVE}: A joint point process model for information diffusion and
  network co-evolution.
\newblock {\em Advances in Neural Information Processing Systems}, 2015.

\bibitem{farajtabar2016msc}
M.~Farajtabar, X.~Ye, S.~Harati, L.~Song, and H.~Zha.
\newblock Multistage campaigning in social networks.
\newblock In {\em Advances in Neural Information Processing Systems}, pages
  4718--4726, 2016.

\bibitem{friggeri2014rumor}
A.~Friggeri, L.~A. Adamic, D.~Eckles, and J.~Cheng.
\newblock Rumor cascades.
\newblock In {\em ICWSM}, 2014.

\bibitem{garrett2013}
C.~Garrett.
\newblock {\em Numerical integration of matrix Riccati differential equations
  with solution singularities}.
\newblock PhD thesis, The University of Texas at Arlington, May 2013.

\bibitem{Rodriguez2011}
M.~Gomez-Rodriguez, D.~Balduzzi, and B.~Sch\"{o}lkopf.
\newblock Uncovering the temporal dynamics of diffusion networks.
\newblock In {\em Proceedings of the 28th International Conference on Machine
  Learning (ICML'11)}, pages 561--568, 2011.

\bibitem{hanson2007}
F.~Hanson.
\newblock {\em Applied stochastic processes and control for Jump-diffusions:
  modeling, analysis, and computation}, volume~13.
\newblock Siam, 2007.

\bibitem{Hawkes1971}
A.~Hawkes.
\newblock Spectra of some self-exciting and mutually exciting point processes.
\newblock {\em Biometrika}, 58(1):83--90, 1971.

\bibitem{smart16}
M.~Karimi, E.~Tavakoli, M.~Farajtabar, L.~Song, and M.~Gomez-Rodriguez.
\newblock {Smart Broadcasting: Do you want to be seen?}
\newblock In {\em Proceedings of the 22nd ACM SIGKDD International Conference
  on Knowledge Discovery in Data Mining}, 2016.

\bibitem{Kempe2003}
D.~Kempe, J.~Kleinberg, and {\'E}.~Tardos.
\newblock Maximizing the spread of influence through a social network.
\newblock In {\em Proceedings of the ninth ACM SIGKDD international conference
  on Knowledge discovery and data mining}, pages 137--146. ACM, 2003.

\bibitem{Kingman1992}
J.~Kingman.
\newblock {\em Poisson processes}.
\newblock Oxford university press, 1992.

\bibitem{Leskovec2010}
J.~Leskovec, D.~Chakrabarti, J.~Kleinberg, C.~Faloutsos, and Z.~Ghahramani.
\newblock Kronecker graphs: An approach to modeling networks.
\newblock {\em The Journal of Machine Learning Research}, 11:985--1042, 2010.

\bibitem{LeskovecCKFG10}
J.~Leskovec, D.~Chakrabarti, J.~M. Kleinberg, C.~Faloutsos, and Z.~Ghahramani.
\newblock Kronecker graphs: An approach to modeling networks.
\newblock {\em JMLR}, 2010.

\bibitem{lewis1979simulation}
P.~A. Lewis and G.~S. Shedler.
\newblock Simulation of nonhomogeneous poisson processes by thinning.
\newblock {\em Naval research logistics quarterly}, 26(3):403--413, 1979.

\bibitem{hdhp2017learning}
C.~Mavroforakis, I.~Valera, and M.~Gomez-Rodriguez.
\newblock Modeling the dynamics of online learning activity.
\newblock In {\em Proceedings of the 26th International World Wide Web
  Conference}, 2017.

\bibitem{Ogata1981}
Y.~Ogata.
\newblock On lewis' simulation method for point processes.
\newblock {\em Information Theory, IEEE Transactions on}, 27(1):23--31, 1981.

\bibitem{reddy2016unbounded}
S.~Reddy, I.~Labutov, S.~Banerjee, and T.~Joachims.
\newblock Unbounded human learning: Optimal scheduling for spaced repetition.
\newblock {\em Proceedings of the 22nd ACM SIGKDD International Conference on
  Knowledge Discovery and Data Mining}, 2016.

\bibitem{RicDom02}
M.~Richardson and P.~Domingos.
\newblock Mining knowledge-sharing sites for viral marketing.
\newblock In {\em Proceedings of the eighth ACM SIGKDD international conference
  on Knowledge discovery and data mining}, pages 61--70. ACM, 2002.

\bibitem{spasojevic2015post}
N.~Spasojevic, Z.~Li, A.~Rao, and P.~Bhattacharyya.
\newblock When-to-post on social networks.
\newblock In {\em Proceedings of the 21th ACM SIGKDD International Conference
  on Knowledge Discovery and Data Mining}, pages 2127--2136. ACM, 2015.

\bibitem{stone1948}
M.~Stone.
\newblock The generalized weierstrass approximation theorem.
\newblock {\em Mathematics Magazine}, 21(5):237--254, 1948.

\bibitem{reliability2017tabibian}
B.~Tabibian, I.~Valera, M.~Farajtabar, L.~Song, B.~Schoelkopf, and
  M.~Gomez-Rodriguez.
\newblock Distilling information reliability and source trustworthiness from
  digital traces.
\newblock In {\em Proceedings of the 26th International World Wide Web
  Conference}, 2017.

\bibitem{Valera2015}
I.~Valera, M.~Gomez-Rodriguez, and K.~Gummadi.
\newblock Modeling diffusion of competing products and conventions in social
  media.
\newblock {\em ICDM}, 2015.

\bibitem{wang2017}
Y.~Wang, E.~Theodorou, A.~Verma, and L.~Song.
\newblock A unifying framework for guiding point processes with stochastic
  intensity functions.
\newblock {\em arXiv preprint arXiv:1701.08585}, 2017.

\bibitem{redqueen17wsdm}
A.~Zarezade, U.~Upadhyay, H.~Rabiee, and M.~Gomez-Rodriguez.
\newblock Redqueen: An online algorithm for smart broadcasting in social
  networks.
\newblock In {\em Proceedings of the 10th ACM International Conference on Web
  Search and Data Mining}, 2017.

\bibitem{zhao2015seismic}
Q.~Zhao, M.~Erdogdu, H.~He, A.~Rajaraman, and J.~Leskovec.
\newblock Seismic: A self-exciting point process model for predicting tweet
  popularity.
\newblock In {\em Proceedings of the 21th ACM SIGKDD International Conference
  on Knowledge Discovery and Data Mining}, 2015.

\end{thebibliography}

\clearpage
\newpage

\begin{appendix}
\label{sec:appendix}
\section{Proof of Proposition \ref{prop:hawkes}}
\label{app:hawkes-dynamics}
Using the left continuity of poisson processes and the definition of derivative $d\bm{\lambda}(t)=\bm{\lambda}(t+dt) - \bm{\lambda}(t)$ we can find the dynamics of the process by the Ito's calculus \cite{hanson2007} as follows:
\begin{align*}
	d\bm{\lambda}(t)
	&= \bm{A} \int_0^{t+dt} g(t+dt-s)\,d\bm{N}(s) - \bm{A} \int_0^t g(t-s)\,d\bm{N}(s) \\
	&=\bm{A} \int_0^{t+dt} (g(t-s)+g'(t-s)dt) \, d\bm{N}(s) - \bm{A} \int_0^t g(t-s)\,d\bm{N}(s) \\
	&=\bm{A} \int_t^{t+dt} g(t-s)\,d\bm{N}(s) + dt \, \bm{A} \int_0^{t+dt} g'(t-s)\,d\bm{N}(s) \\
	&=\bm{A}\,g(0) \, d\bm{N}(t) - w \, dt \, \bm{A} \int_0^{t+dt} g(t-s)\,d\bm{N}(s)	 \\
	&=\bm{A}\,d\bm{N}(t)  - w \, dt \, \bm{A} \int_0^{t} g(t-s)\,d\bm{N}(s)		 \\
	&=\bm{A}\,d\bm{N}(t) + w\,dt\,[\bm{\mu}_0(t)-\bm{\lambda}(t)]
\end{align*}

\section{Proof of Theorem \ref{thm:bellman-opt-cond}}
\label{app:bellman-opt-cond}
\begin{align*}
	&J(\bm{\lambda}(t),t) =
	\min_{\bm{u}(t,t_f]} \mathbb{E}_{(\bm{N},\bm{M})(t,t_f]} \left[ \phi(\bm{\lambda}(t_f)) + \int_t^{t_f} \ell(\bm{\lambda}(s),\bm{u}(s)) \, ds \right ] \\
	&= \min_{\bm{u}(t,t_f]} \mathbb{E}_{(\bm{N},\bm{M})(t,t_f]} \bigg[ \phi(\bm{\lambda}(t_f)) + \int_t^{t+dt} \ell(\bm{\lambda}(s),\bm{u}(s)) \, ds 
	 + \int_{t+dt}^{t_f} \ell(\bm{\lambda}(s),\bm{u}(s)) \, ds \bigg ]  \\
	&= \min_{\bm{u}(t,t_f]} \mathbb{E}_{(\bm{N},\bm{M})(t,t+dt]}\bigg[\mathbb{E}_{(\bm{N},\bm{M})(t+dt,t_f]} \Big[ \phi(\bm{\lambda}(t_f)) + \ell(\bm{\lambda}(t),\bm{u}(t)) \, dt  
	+ \int_{t+dt}^{t_f} \ell(\bm{\lambda}(s),\bm{u}(s)) \, ds \Big ] \bigg] \\
	&= \min_{\bm{u}(t,t+dt]} \min_{\bm{u}(t+dt,t_f]} \mathbb{E}_{(\bm{N},\bm{M})(t,t+dt]}\bigg[\ell(\bm{\lambda}(t),\lambda(t),t) \, dt 
	+\mathbb{E}_{(\bm{N},\bm{M})(t+dt,t_f]} \Big[ \phi(\bm{\lambda}(t_f)) + \int_{t+dt}^{t_f} \ell(\bm{\lambda}(s),\bm{u}(s)) \, ds \Big ] \bigg]  \\
	&= \min_{\bm{u}(t,t+dt]} \mathbb{E}_{(\bm{N},\bm{M})(t,t+dt]} \left[ J(\bm{\lambda}(t+dt),t+dt)\right ]
	+ \ell(\bm{\lambda}(t),\bm{u}(t)) \, dt
\end{align*}

\section{Proof of Theorem \ref{thm:activity-diff-cost}}
\label{app:activity-diff-cost}
Using the definition of derivative we can evaluate the differential of the cost-to-go as follows:
\begin{align*}
	dJ(\bm{\lambda}(t),t) &= J(\bm{\lambda}(t+dt),t+dt) - J(\bm{\lambda}(t),t) \\
	&=J(\bm{\lambda}(t)+d\bm{\lambda}(t),t+dt) - J(\bm{\lambda}(t),t).
\end{align*}
To evaluate the first term in the right hand side of the above equality we substitute $d\bm{\lambda}(t)$ by the SDE dynamics (\ref{eq:sys-dyn}). Then, using the zero-one jump law~\cite{Kingman1992} we can write:
\begin{align*}
	&J(\bm{\lambda}(t)+d\bm{\lambda}(t),t+dt) = J(\bm{\lambda}(t)+ f(t)\,dt + \bm{A} \, d\bm{N}(t) + \bm{A} \, d\bm{M}(t), t+dt) \\
	&= \sum_i J(\bm{\lambda}(t)+ f(t)\,dt + \bm{a}_i , t+dt) dN_i(t) + \sum_i J(\bm{\lambda}(t)+ f(t)\,dt + \bm{a}_i, t+dt) dM_i(t) \\
	&\quad+ J(\bm{\lambda}(t)+ f(t)\,dt, t+dt) \prod_i [1-dN_i(t)][1-dM_i(t)] \\
	&=J(\bm{\lambda}(t)+ f(t)\,dt, t+dt) [1-\sum_{i} dN_i(t)+dM_i(t)] + \sum_i J(\bm{\lambda}(t)+ f(t)\,dt + \bm{a}_i , t+dt) [dN_i(t)+dM_i(t)] 
	 \\
	&=J(\bm{\lambda}(t)+ f(t)\,dt, t+dt) + \sum_i [J(\bm{\lambda}(t)+ f(t)\,dt + \bm{a}_i , t+dt) - J(\bm{\lambda}(t)+ f(t)\,dt, t+dt)] [dN_i(t)+dM_i(t)] 
\end{align*}
where we denote $f(t) := w \bm{\mu}_0 - w \bm{\lambda}(t)$ for notation simplicity, and used  that the bilinear differential form $dt \, dN(t)=0$ as in~\cite{hanson2007}.
By total derivative rule we have:
\begin{align*}
	J(\bm{\lambda}(t)+ f(t)\,dt + \bm{a}_i , t+dt) &=
	J(\bm{\lambda}(t)+\bm{a}_i,t) + \nabla_{\bm{\lambda}}J(\bm{\lambda}(t)+\bm{a}_i,t) f(t)\,dt + J_t(\bm{\lambda}(t)+\bm{a}_i,t)\,dt \\
	J(\bm{\lambda}(t)+ f(t)\,dt, t+dt) &=
	J(\bm{\lambda}(t),t) + \nabla_{\bm{\lambda}}J(\bm{\lambda}(t),t) f(t)\,dt + J_t(\bm{\lambda}(t),t)\,dt. 	
\end{align*}
If we plug these two relation in to the previous one, it results to:
\begin{align*}
	J(\bm{\lambda}(t)+d\bm{\lambda}(t),t+dt) &=  J(\bm{\lambda}(t),t) + \nabla_{\bm{\lambda}}J(\bm{\lambda}(t),t) f(t)\,dt + J_t(\bm{\lambda}(t),t)\,dt \\
	&+ \sum_i [J(\bm{\lambda}(t)+\bm{a}_i , t) - J(\bm{\lambda}(t), t)] [dN_i(t)+dM_i(t)],
\end{align*}
which completes the proof.

\section{Proof of $(\Delta_A J)_i \leq 0$} 
\label{app:positive-delta}
Lets $t < s$, then according to the definition we can write,
\begin{align*}
	\bm{\lambda}(s) = \bm{\mu}_0 + \int_0^s g(s-\tau) \, \bm{A} \, d\bm{N}(\tau) = \bm{\mu}_0 + \int_0^t g(s-\tau) \, \bm{A} \, d\bm{N}(\tau) + \int_t^s g(s-\tau) \, \bm{A} \, d\bm{N}(\tau).
\end{align*}
For the exponential kernel $g(t)=e^{-wt}$ we have,
\begin{align*}
	\int_0^t g(s-\tau) \, \bm{A} \, d\bm{N}(\tau) = \int_0^t e^{-w(s-\tau)} \, \bm{A} \, d\bm{N}(\tau) = e^{-w(s-t)} \int_0^t e^{-w(t-\tau)} \, \bm{A} \, d\bm{N}(\tau) = e^{-w(s-t)} (\bm{\lambda}(t) - \bm{\mu}_0 )
\end{align*}
so given the value of $\bm{\lambda}(t)$ at time $t$ then we can write $\bm{\lambda}(s)$ for later times as
\begin{align*}
	\bm{\lambda}(s) &= \bm{\mu}_0 + e^{-w(s-t)} (\bm{\lambda}(t) - \bm{\mu}_0 )
	 + \int_t^s g(s-\tau) \, \bm{A} \, d\bm{N}(\tau)
\end{align*}
Lets consider a process $\bm{\xi}(s)$ with intensity value at time $t$ equal to $\bm{\lambda}(t)+\bm{a}_i$ as,
\begin{align*}
	\bm{\xi}(s) &= \bm{\mu}_0 + e^{-w(s-t)} (\bm{\lambda}(t) + \bm{a}_i - \bm{\mu}_0 )
	 + \int_t^s g(s-\tau) \, \bm{A} \, d\bm{N}(\tau).
\end{align*}
Since $\bm{a}_i \succeq 0$, then given the same history in interval $(t,s)$, we have $\bm{\xi}(s) \succeq \bm{\lambda}(s)$. 
Then, we have:
\begin{align*}
	\ell(\bm{\xi}(s), \bm{u}(s)) \leq \ell(\bm{\lambda}(s), \bm{u}(s)).
\end{align*}

Now, taking the integration, then expectation (over all histories) and finally the minimization from the above inequality does not change the direction of inequality. So it readily follows the required result.  
%
%
\begin{align*}
	J(\bm{\lambda}(t)+\bm{a}_i, t) \leq J(\bm{\lambda}(t), t)
\end{align*}

\section{Proof of Lemma \ref{lem:quad-proposal}} \label{sec:quad-proposal} 
Consider the following proposal of degree three for the cost-to-go function:
\begin{align*}
	J(\bm{\lambda}(t), t) = f(t) + \sum_i g_i(t)  \lambda_i(t) + \sum_i \sum_j \lambda_i(t) \lambda_j(t) H_{ij}(t) + \sum_i\sum_j\sum_k \lambda_i(t) \lambda_j(t) \lambda_k(t) H_{ijk}(t)
\end{align*}
If we plug this proposal in Eq.~\ref{eq:hjb}, and evaluate the coefficient of fourth degree terms like $ \lambda_i^2(t)\lambda_j^2(t)$ and equate them to zero, then we can find the unknown coefficients $H_{ijk}(t)$'s as follows:
\begin{align*}
	\forall i,j,t: \,\,	\sum_k \big(\sum_\ell A_{\ell k}^2 \big) H^2_{ijk}(t) = 0
\end{align*}
Since the sum of positives terms is zero if and only if they all be zero, then $H_{ijk}(t)$ and consequently the terms with degree three in the proposal are all zero. So the proposal reduces to a quadratic proposal. 

It is quite straightforward to extend this argument for proposals with order $m > 3$  and by equating the degree $2m-1$ terms, similarly conclude that coefficients of degree $m$ terms in the proposal are zero. If we repeat this argument for $m-1,\ldots,3$, we deduce that any proposal with arbitrary degree $m \geq 2$, would result in a quadratic optimal cost. 

Finally, according to the Stone-Weierstrass theorem, any continuous function in a closed interval can be approximated as closely as desired by a polynomial function \cite{stone1948}. So by assuming the continuity of cost function, it can be approximate by a polynomial as closely as desired, and then it also reduce to a quadratic proposal too.
\begin{figure*}[t]
\centering
  { \includegraphics[width=0.956\textwidth]{legend_IwB_new}}\\ \vspace*{-3mm}
 \hspace*{-0.6cm}\subfloat[Core-periphery]{\setcounter{subfigure}{1} \includegraphics[height=0.18\textwidth]{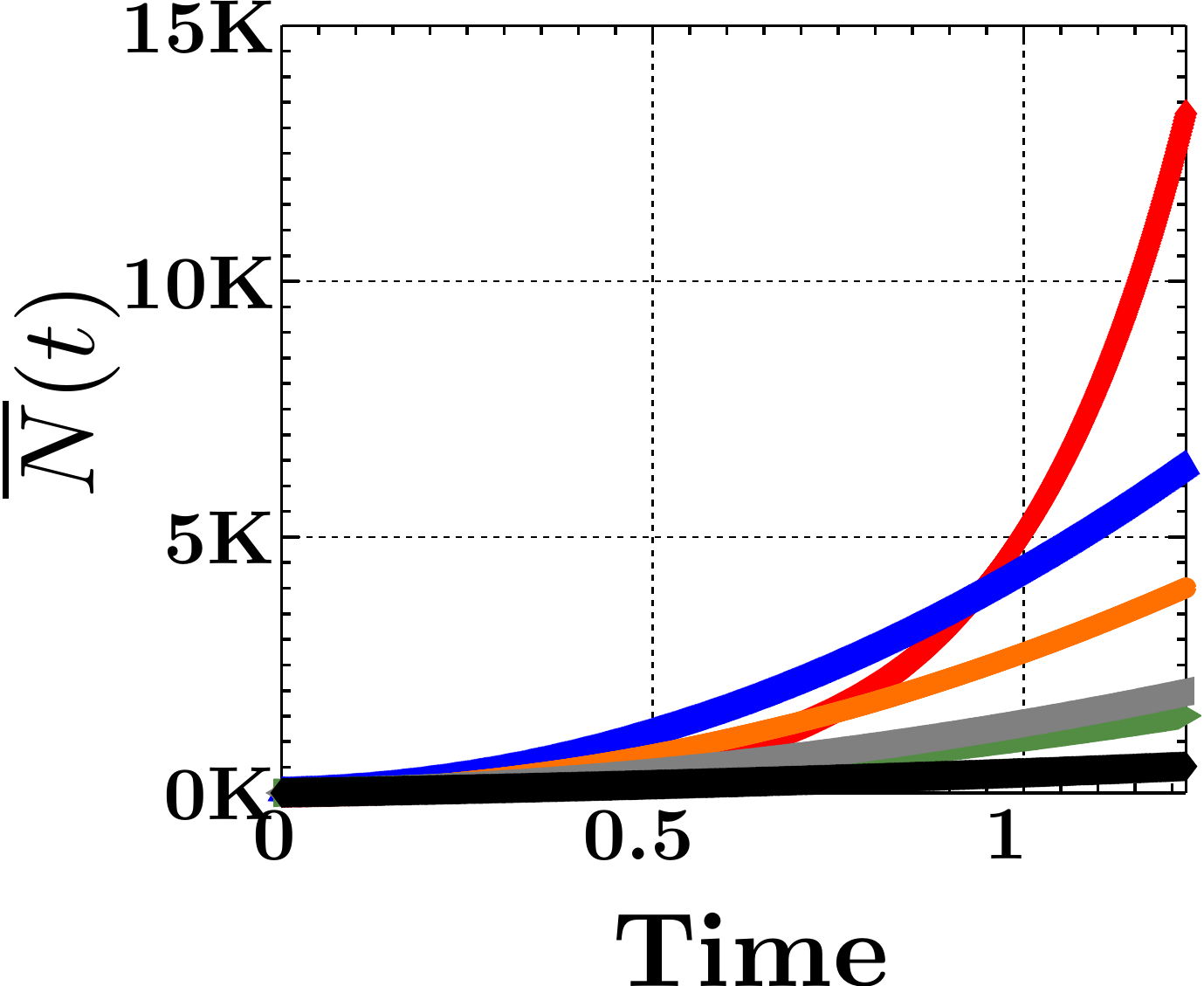}}
\hspace*{0.0cm}\subfloat[Hierarchical]{\includegraphics[height=0.18\textwidth]{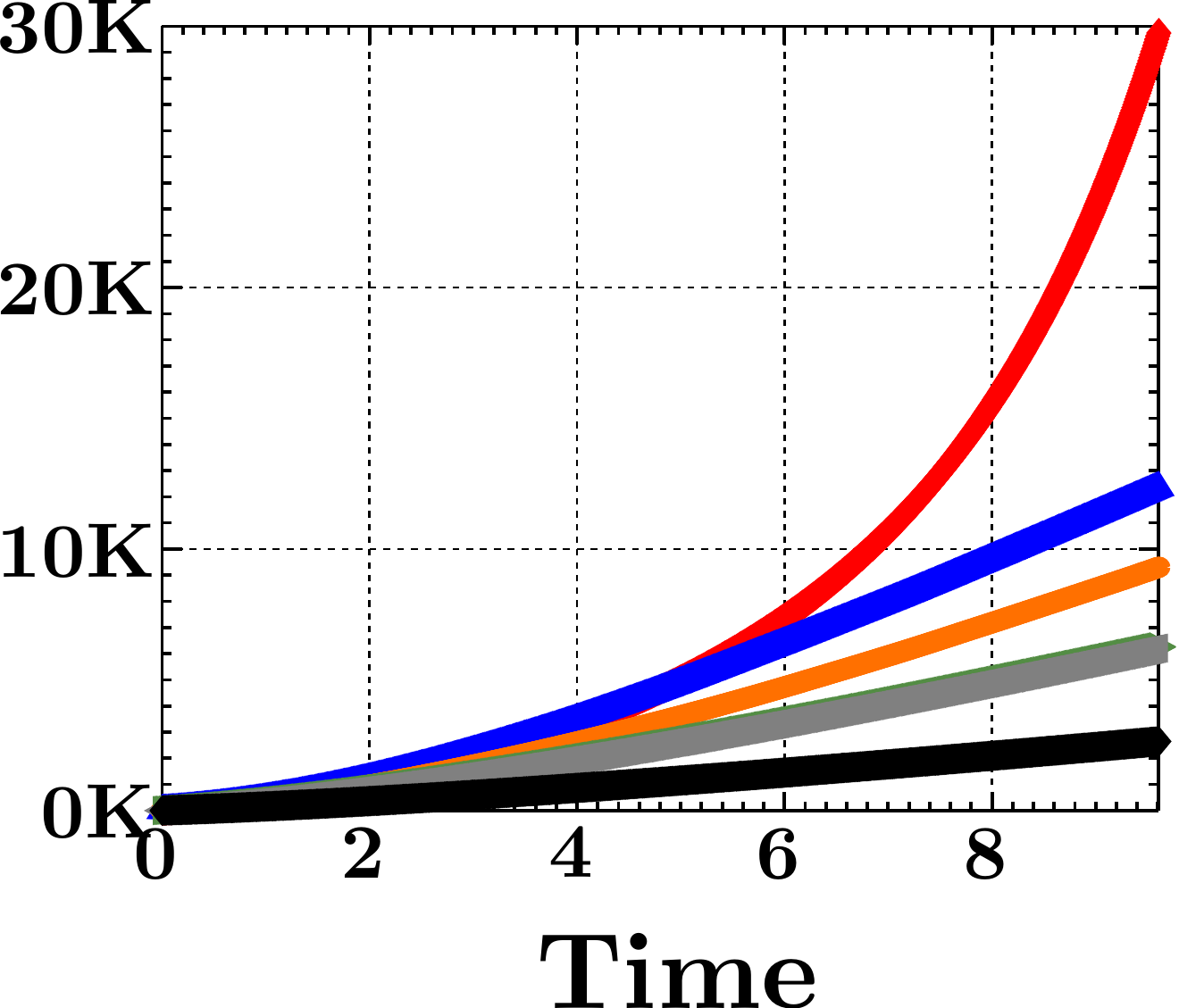}}
\hspace*{0.0cm}\subfloat[Homophily]{\includegraphics[height=0.18\textwidth]{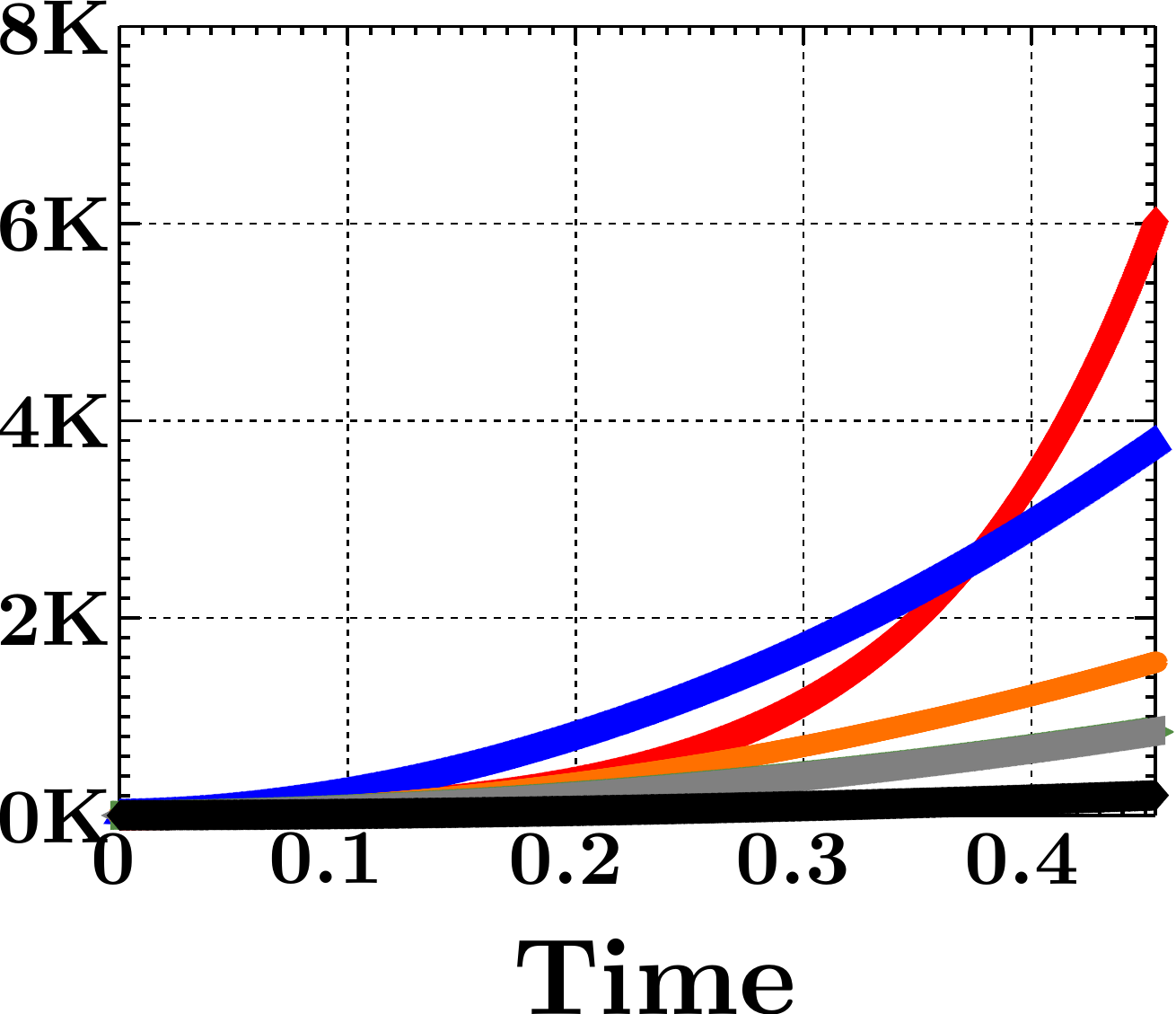}}  
\hspace*{0.0cm}\subfloat[Heterophily]{ \includegraphics[height=0.18\textwidth]{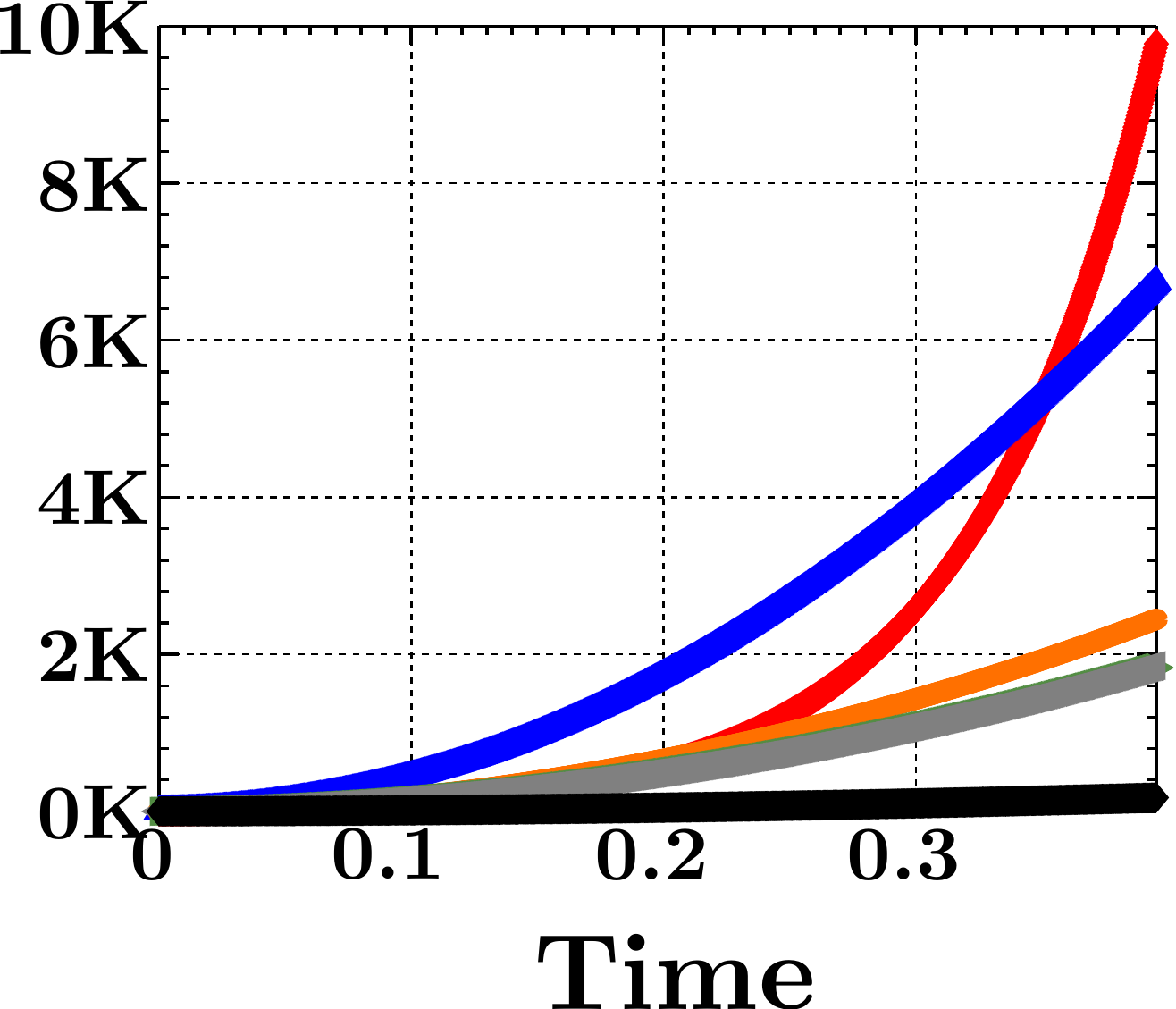}}
\hspace*{0.0cm}\subfloat[Random]{ \includegraphics[height=0.18\textwidth]{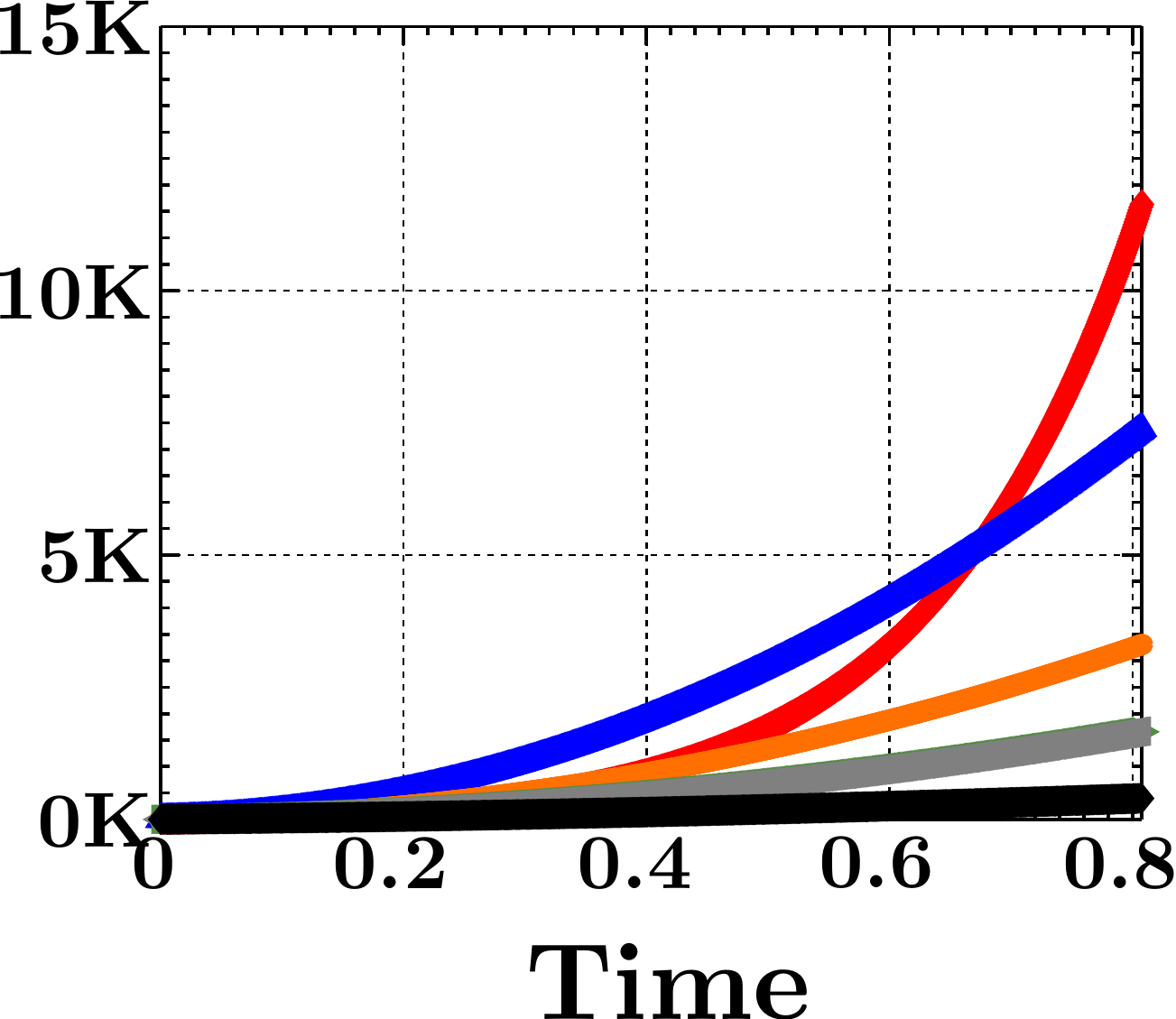}}
\vspace{-1mm}
\caption{Performance over time of \cheshire against several competitors for several types of Kronecker networks. Performance is measured in terms of overall 
number of tweets $\bar{N}(t) = \sum_{u \in \Vcal} \EE[N_u(t)]$. In all cases, we tune the parameters $Q$, $S$ and $F$ such that the total number of incentivized 
tweets \emph{posted} by our method is equal to the budget used in the competing methods and baselines.}
\label{fig:NwTSyn}
\vspace{-2mm}
\end{figure*}

\section{Additional Experiments on Synthetic Data} \label{app:synthetic}
\xhdr{Experimental setup}
In this section, we experiment with five different types of Kronecker networks~\cite{LeskovecCKFG10} with $512$ nodes:
(i) assortative networks (parameter matrix $[0.96 , 0.3; 0.3, 0.96]$); 
(ii) dissortative networks ($[0.3, 0.96 ; 0.96,0.3]$); 
(iii) random networks ($[0.7, 0.7; 0.7,0.7]$);
(iv) hierarchical networks ($[0.9, 0.1 ; 0.1, 0.9]$); and,
(v) core-periphery networks ($[0.9, 0.5 ; 0.5, 0.3]$).
For each network, we draw $\mub$ and $\Ab$ from a uniform distribution $U(0,1)$ and set $\omega=100$.
Similarly as in the experiments with Twitter data, we compare the performance of our algorithm with two state of the art methods, OPL~\cite{Farajtabar2015} and 
MSC~\cite{farajtabar2016msc}, and three baselines, PRK, DEG and UNC. 

\xhdr{Performance}
Figure~\ref{fig:NwTSyn} compares the performance of our algorithm against others in terms of overall average number of tweets $\bar{N}(t) = \sum_{u \in \Vcal} \EE[N_u(t)]$ for a 
fixed budget $\bar{M}(t_f) = \sum_{u \in \Vcal} M_u(t_f) \approx 6.1$K. We find that: (i) our algorithm consistently outperforms competing methods by large margins at time $t_f$; (ii) it triggers up 
to $50$\%--$100$\% more posts than the second best performer by time $t_f$; (iii) MSC tends to use the budget too early, as a consequence, although it initially beats our method,
it eventually gets outperformed by time $t_f$; and, (iv) the baselines PRK and DEG have an underwhelming performance, suggesting that the network structure alone is not an 
accurate measure of influence.

\xhdr{Scalability} 
Figure~\ref{fig:NwTSyn} shows that our algorithm scales to large networks and is almost an order of magnitude faster than the second best 
performer, MSC~\cite{farajtabar2016msc}. For example, our algorithm takes $\sim$$30$ seconds to steer a network with $1{,}000$ nodes and 
average degree of $10$ while MSC takes  $\sim$$4$ minutes.

\begin{figure*}[t]
\centering
 \hspace*{-0.6cm}\subfloat[Running time vs $t_f$]{\setcounter{subfigure}{1} \includegraphics[height=0.18\textwidth]{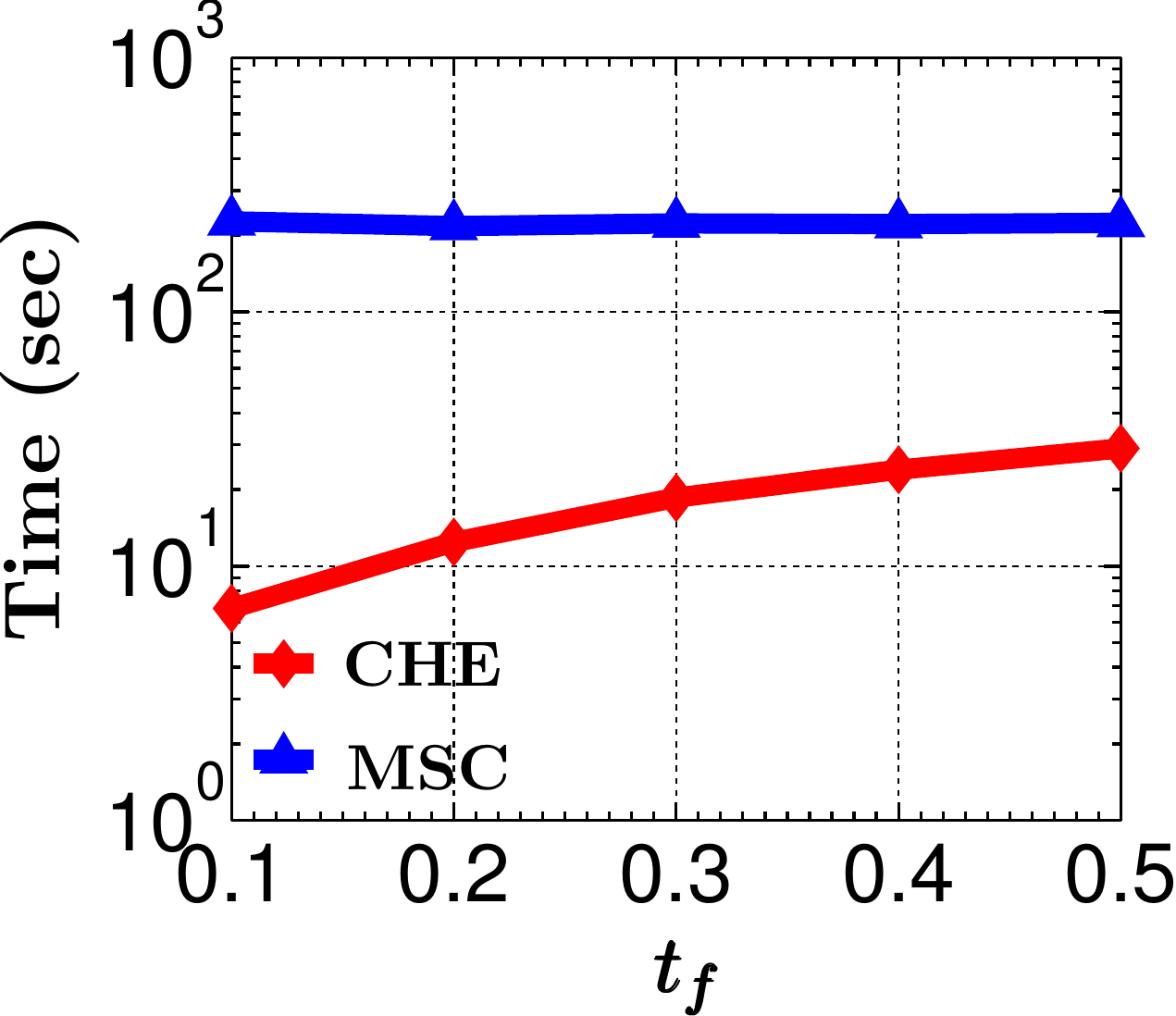}}
\hspace*{1cm}\subfloat[Running time vs \#nodes]{\includegraphics[height=0.18\textwidth]{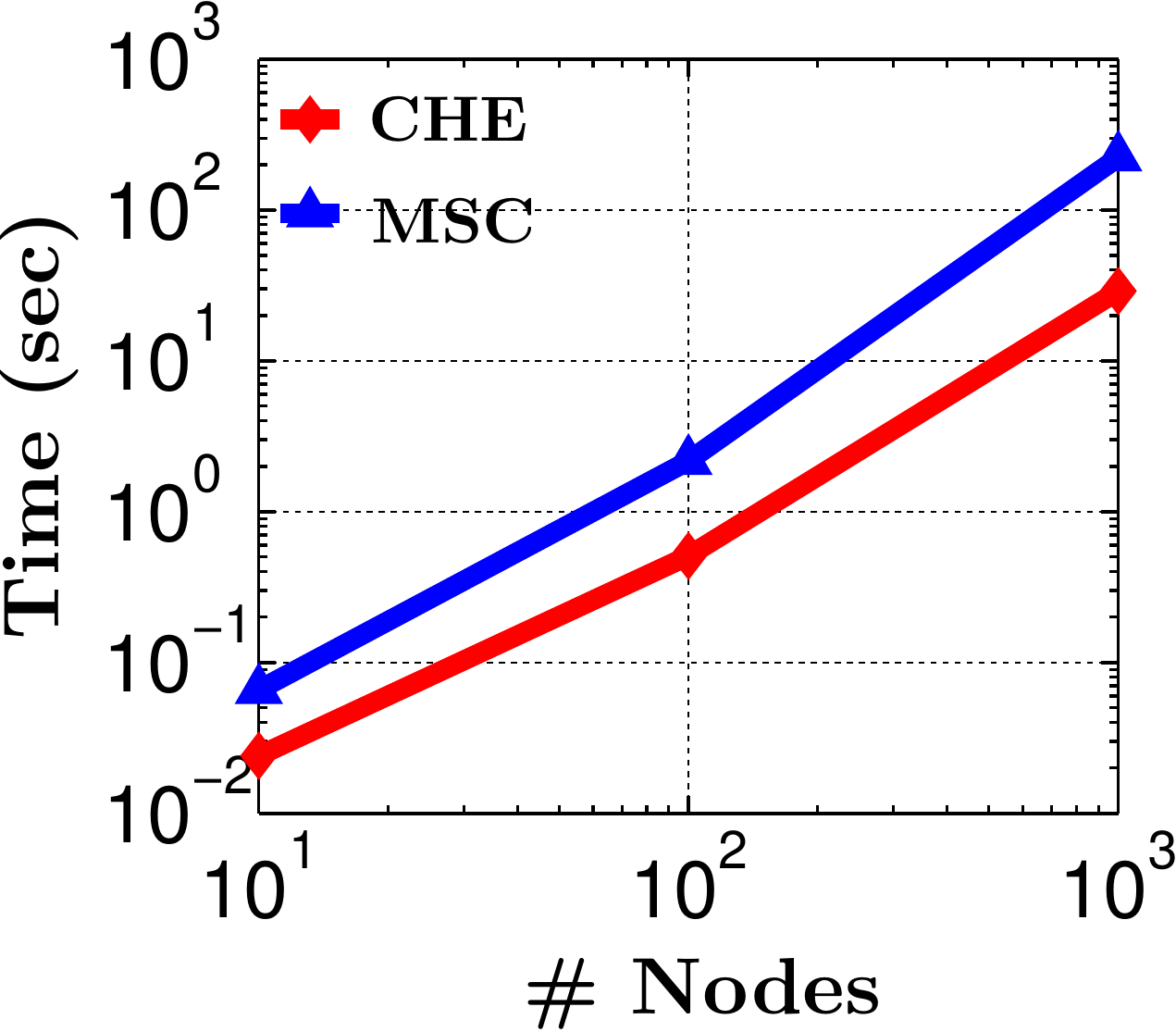}}
 \vspace{-1mm}
\caption{Scalability of \cheshire against several competitors. Panel (a) shows the running time against the cut-off time $t_f$ for a $1{,}000$
node Kronecker network. Panel (b) shows the running time for several Kronecker networks of increasing size with $t_f = 0.5$. In both panels, 
the average degree per node is $10$. The experiments are carried out in a single machine with 24 cores and 64 GB of main memory.}
\label{fig:NwTSyn}
\vspace{-2mm}
\end{figure*}

\section{Twitter Datasets Description} \label{app:real-datasets}
We used the Twitter search API\footnote{\scriptsize \url{https://dev.twitter.com/rest/public/search}} to collect all the tweets (corresponding to a 2-3 weeks period 
around the event date) that contain hashtags related to the following events/topics: 
\begin{itemize}[noitemsep,nolistsep]
\item \textbf{Elections: } British election, from May 7 to May 15, 2015.
\item \textbf{Verdict:} Verdict for the corruption-case against Jayalalitha, an Indian politician, from May 6 to May 17, 2015.
\item \textbf{Club:} Barcelona getting the first place in La-liga, from May 8 to May 16, 2016. 
\item \textbf{Sports: } Champions League final in 2015, between Juventus and Real Madrid, from May 8 to May 16, 2015.
\item \textbf{TV Show:} The promotion on the TV show ``Games of Thrones'', from May 4 to May 12, 2015. 
\end{itemize}
We then built the follower-followee network for the users that posted the collected tweets using the Twitter rest API\footnote{\scriptsize \url{https://dev.twitter.com/rest/public}}. 
Finally, we filtered out users that posted less than 200 tweets during the account lifetime, follow less than 100 users, or have less than 50 followers. An account of the dataset statistics is given in Table~\ref{tab:datasets}.\\
\begin{table}[!h]
\small
{\centering\begin{tabular}{|p{2.4cm}|c|c|c|c|}
\hline%
 \textbf{Dataset}&\textbf{  $\mathbf{|\Vcal|}$} & \textbf{$\mathbf{|\Ecal|}$} & \textbf{$|\Hcal(T_{\text{Data}})|$} &\textbf{$T=T_\text{simulation}$} \\ \hline \hline
{Elections}  &   231 &1108  	& 1584 	& 120.2\\ \hline
{Verdict}    &  1059  & 10691  & 17452	& 22.11 \\ \hline
{Club}       &  703  & 4154  	&  9409	& 19.23\\ \hline
{Sports}      &  703  & 4154    &  7431	& 21.53\\ \hline%
{TV Show}      &  947 & 10253 	& 13203	& 12.11\\ \hline%
\end{tabular}
\caption{Real datasets statistics}
\label{tab:datasets}}
\end{table}

\end{appendix}

\end{document}